\documentclass[journal,twoside,web]{ieeecolor2}
\usepackage{generic}
\usepackage{cite}
\usepackage{amsmath,amssymb,amsfonts}
\usepackage{algorithm}
\usepackage{algorithmic}
\usepackage{graphicx}
\usepackage{textcomp}
\usepackage{hyperref}
\usepackage{placeins}
\usepackage{booktabs}

\usepackage{graphics} 
\usepackage{epsfig} 
\usepackage{mathptmx} 
\usepackage{url}
\usepackage{soul}

\def\BibTeX{{\rm B\kern-.05em{\sc i\kern-.025em b}\kern-.08em
    T\kern-.1667em\lower.7ex\hbox{E}\kern-.125emX}}
\begin{document}
\title{Robust Neural Stimulation Response Modeling Through Meta-Learning and Pretraining}
\author{Matthew J Bryan$^{1,2,3}$, Daniel C Muir$^{1,2,3}$, Felix Schwock$^{2,3,5}$, Azadeh Yazdan-Shahmorad$^{2,3,4,5}$, Rajesh P N Rao$^{1,2,3}$%
\thanks{This work was supported by National Science Foundation (NSF) EFRI
grant no.\ 2223495, a Weill Neurohub Investigator grant, a CJ and Elizabeth Hwang endowed professorship (RPNR), National Institutes of Health grant nos.\ R01NS119593, R01MH125429 (FS, AYS), and the National Defense Science and Engineering Graduate (NDSEG) Fellowship Program NDSEG10578COMPSCI (MJB). This material is based upon work supported by the Air Force Office of Scientific Research under award number FA9550-23-F-0014 in the amount of \$139,900 USD (MJB). Any opinions, findings, and conclusions or recommendations expressed in this material are those of the authors and do not necessarily reflect the views of the funders.}
\thanks{Author emails (respectively): mmattb@cs.washington.edu, danmuir@cs.washington.edu, fschwock@uw.edu, azadehy@uw.edu, rao@cs.washington.edu}
\thanks{$^{1}$ Neural Systems Laboratory, Paul G. Allen School of Computer Science \& Engineering, University of Washington, Seattle, WA, USA}%
\thanks{$^{2}$ Center for Neurotechnology, University of Washington, Seattle, WA, USA}%
\thanks{$^{3}$ Computational Neuroscience Center, University of Washington, Seattle, WA, USA}%
\thanks{$^{4}$ Department of Bioengineering, University of Washington, Seattle, WA, USA}%
\thanks{$^{5}$ Department of Electrical and Computer Engineering, University of Washington, Seattle, WA, USA}%
}

\maketitle
\thispagestyle{empty}
\pagestyle{empty}

\begin{abstract}

\textbf{Objective:} Model-based closed-loop neural stimulation
holds promise for therapeutic applications ranging from Parkinson's
disease to sensory restoration, but deployment has been limited
by two obstacles: (1) forecasting models for predicting the consequences of stimulation fail catastrophically on a
meaningful fraction of sessions, and (2) per-session calibration
requirements are often incompatible with clinical constraints.
We address both by demonstrating, for the first time, that
meta-learning and pretraining can be applied to neural stimulation response modeling.
\textbf{Methods:} Temporal basis function models (TBFMs) have previously been used to forecast
state-dependent neural responses to stimulation. We extend the application of TBFMs
with cross-session pretraining using a novel architecture and algorithm
based on model-agnostic meta-learning
(MAML), and evaluate the approach on 40 sessions of optogenetic stimulation
in primary somatosensory and motor cortex of two non-human primates.
\textbf{Results:} Meta-learning substantially reduces catastrophic
forecast failure: for a 1k calibration set size, sessions with
test $R^2 < 0.05$ drop from 16/40 (single-session training) to 1/40
(MAML-pretrained), and prediction intervals become significantly
narrower (Brown-Forsythe $p < 0.05$). Calibration requirements
are reduced by 50--90\% at matched accuracy, enabling experiments
otherwise infeasible within clinical session-time constraints.
\textbf{Conclusion:} Our results demonstrate that cross-session structure in stimulation
responses is consistent enough to support pretraining, providing
the first empirical evidence that meta-learning approaches
are viable for neural stimulation.
\textbf{Significance:} The robustness and sample efficiency gains
directly address known obstacles to deploying model-based stimulation
controllers. Our results motivate community efforts to assemble
standardized multi-site stimulation datasets and to further explore
meta-learning for robust closed-loop stimulation.
\end{abstract}


\section{Introduction}
\label{sec:introduction}

\IEEEPARstart{C}{losed}-loop neural stimulation remains underexplored despite its potential for greater efficacy, higher energy efficiency \cite{castano.pd}, and reduced side effects \cite{little.park} compared to open-loop stimulation. It uses sensed brain activity and potentially data from other sensors to adapt stimulation in real time (e.g. \cite{bryan.coproc, bolus.opto}), allowing it to take advantage of ``state-dependence'' \cite{bradley.statedep}, i.e., the dependence of the stimulation response on neural or behavioral states. Clinical open-loop systems have been successfully demonstrated for sensory function restoration, symptom alleviation and targeted rehabilitation for a variety of neurological conditions, e.g., cochlear implants for deafness \cite{niparko.cochlear} to deep brain stimulation (DBS) for Parkinson's disease (PD) \cite{castano.pd}. Closed-loop stimulation seeks to improve on these, but remains difficult due to myriad technical challenges including controller latency, precise characterization and modeling of neural responses to stimulation, and reliable calibration \cite{bryan.embc}.

Adaptive closed-loop stimulation guided by artificial neural networks (ANN) — often termed neural co-processing \cite{bryan.coproc, rao.coproc} - offers a path toward stimulation controllers that adjust dynamically to a patient’s neural state, individual differences, and disease-specific features. Realizing such a neural co-processor for clinical use however will require significant improvements in robust, sample efficient, accurate, and low-latency model-based stimulation controllers. 

Temporal basis function models (TBFMs) \cite{bryan.embc} seek to address these challenges by providing single-trial, multi-step, spatiotemporal forecasts of the state-dependent neural response to stimulation, while exhibiting high sample efficiency and low latency. They improve on the prior state-of-the-art which, for example, did not consider state-dependence \cite{shanechi.stimmodel}, did not operate with low enough latency for control, or offer sufficient sample efficiency for practical use in clinical settings \cite{bryan.coproc}. TBFMs ameliorate these issues: Bryan et al. \cite{bryan.embc} demonstrated their potential to be used in the span of a 1-2 hour experimental session, including data collection for calibration and model training time.

In this article, we advance TBFMs by showing that pretraining on data from past sessions and using model-agnostic meta-learning (MAML) \cite{maml} substantially improves the robustness of stimulation response forecasting. Pretrained TBFMs exhibit significantly fewer sessions where accuracy degrades catastrophically due to noise or other nuisance variables, addressing a known obstacle to deploying model-based stimulation controllers in clinical and experimental settings. The same pretraining approach additionally allows TBFMs to adapt to future sessions with calibration datasets $50-90\%$ smaller than those needed for single-session TBFMs, enabling experiments otherwise infeasible within single session-time constraints.

These gains are notable given that TBFMs were previously demonstrated to have sample efficiency comparable to basic linear state-space models while exhibiting significantly higher prediction accuracy \cite{bryan.embc}. The pretraining and meta-learning approach we introduce here allows experimenters to (1) take advantage of previously collected stimulation data, (2) reduce the frequency of failed experimental sessions and (3) concentrate significantly more experiment time on exploring stimulation parameters in a closed-loop fashion rather than on calibration.

Meta-learning involves the pretraining of models which are easily adapted to unseen tasks \cite{maml}, in our case stimulation experiment sessions. In the case of MAML specifically, model parameters are pushed into regimes where adaptation to an unseen session can be done with minimal training data. The adaptation process is referred to as ``test-time adaptation'' (TTA), and involves fine-tuning some part of the model for the unseen session using a small calibration data set. This pretraining improves robustness by biasing per-session models towards patterns identified across many previous sessions, anchoring the learning process and reducing the degenerate effects of nuisance variables. It has been applied recently to brain-computer interfacing (BCI) to minimize calibration data collection \cite{maml.bci}, but to our knowledge, neither MAML nor meta-learning more generally have previously been shown to be effective for neural stimulation modeling.

Meta-learning works best when there is similarity between sessions which pretraining can leverage. With varying numbers of usable electrodes, differing subjects and electrode placements, and varying stimulation timings and locations, there are many sources of dissimilarity between sessions. Yet local stimulation responses tend to be stereotyped, some spatial autocorrelation statistics may be similar (see example in Fig \ref{fig:spatiotemporal}), and similarly for some temporal autocorrelations \cite{bryan.aacf}. We show for the first time that pretraining and meta-learning can take advantage of these similarities, and that its adaptation process can overcome the cross-session dissimilarities, allowing the model to adapt to unseen sessions while exhibiting accuracy and robustness superior to single-session models. 

In summary, this article provides:
\begin{itemize}
    \item A brief introduction to TBFMs, based on prior work;
    \item A novel machine learning architecture based on TBFMs which allows for cross-session pretraining;
    \item A meta-learning algorithm for pretraining such models;
    \item A test-time adaptation algorithm for unseen sessions;
    \item A free-to-use open source implementation;
    \item Results demonstrating: \begin{enumerate}
        \item A significant improvement in robustness, defined as the the model's accuracy on the highest noise sessions;
        \item Significant savings in calibration data collection; and
        \item Comparable or improved forecast accuracy relative to single-session models across all calibration data set sizes.
    \end{enumerate}
\end{itemize}

\section{Methods}
\label{sec:methods}
\subsection{Dataset}
\begin{figure}[thpb]
  \centering
  \includegraphics[width=\linewidth]{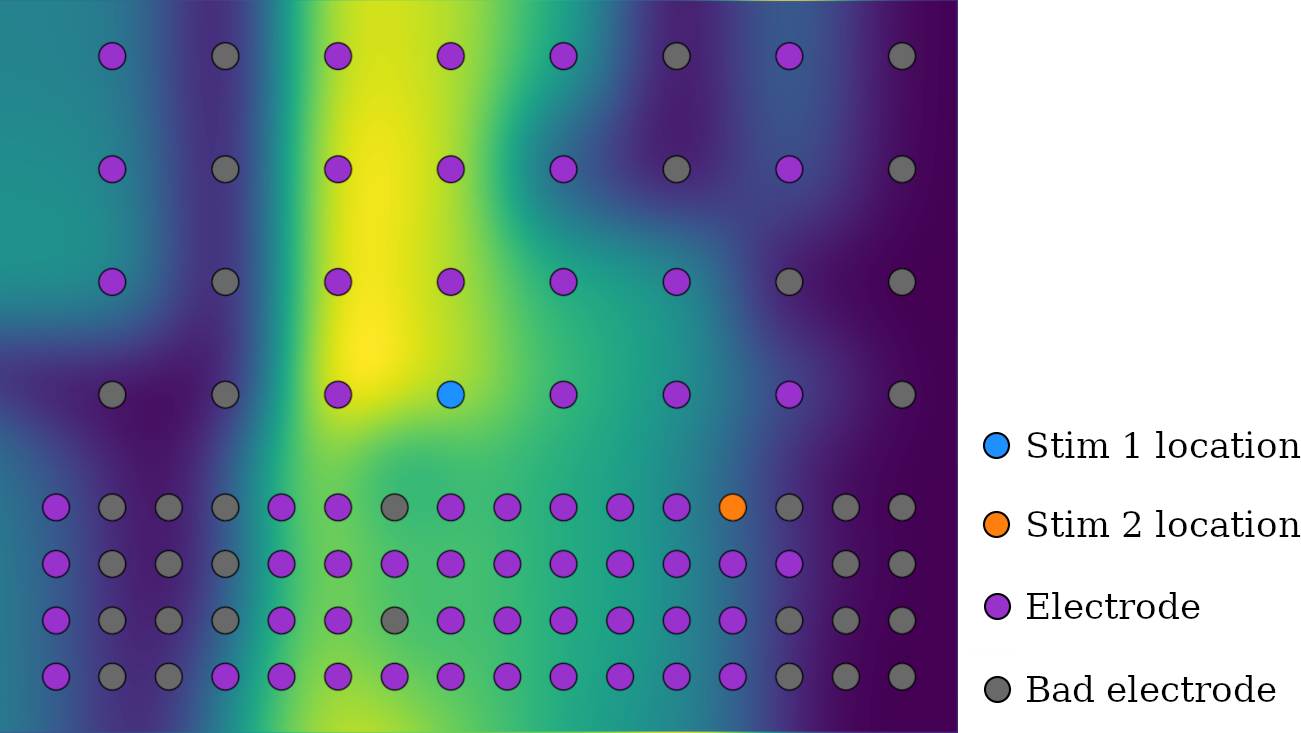}
  \caption{\textbf{Stimulation response across the $\mu$ECog array.} Heatmap indicates LFP values (Gaussian-smoothed and Z-scored, single session average) at 10ms after the second stim pulse onset. Missing electrodes were given false values for graphing. Stim locations varied between sessions.}
  \label{fig:spatiotemporal}
\end{figure}

\begin{figure}[thpb]
  \centering
  \includegraphics[width=\linewidth]{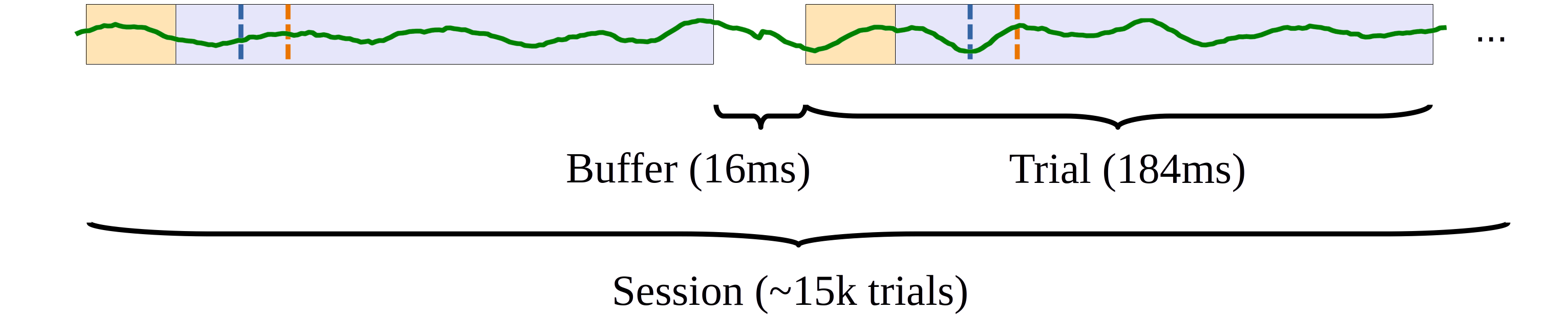}
  \caption{\textbf{Data windowing} Trials are 184ms separated by a 16ms buffer, $\sim15k$ trials per session. See Figure~\ref{fig:trial} for legend.}
  \label{fig:timeline}
\end{figure}

\begin{figure}[thpb]
  \centering
  \includegraphics[width=\linewidth]{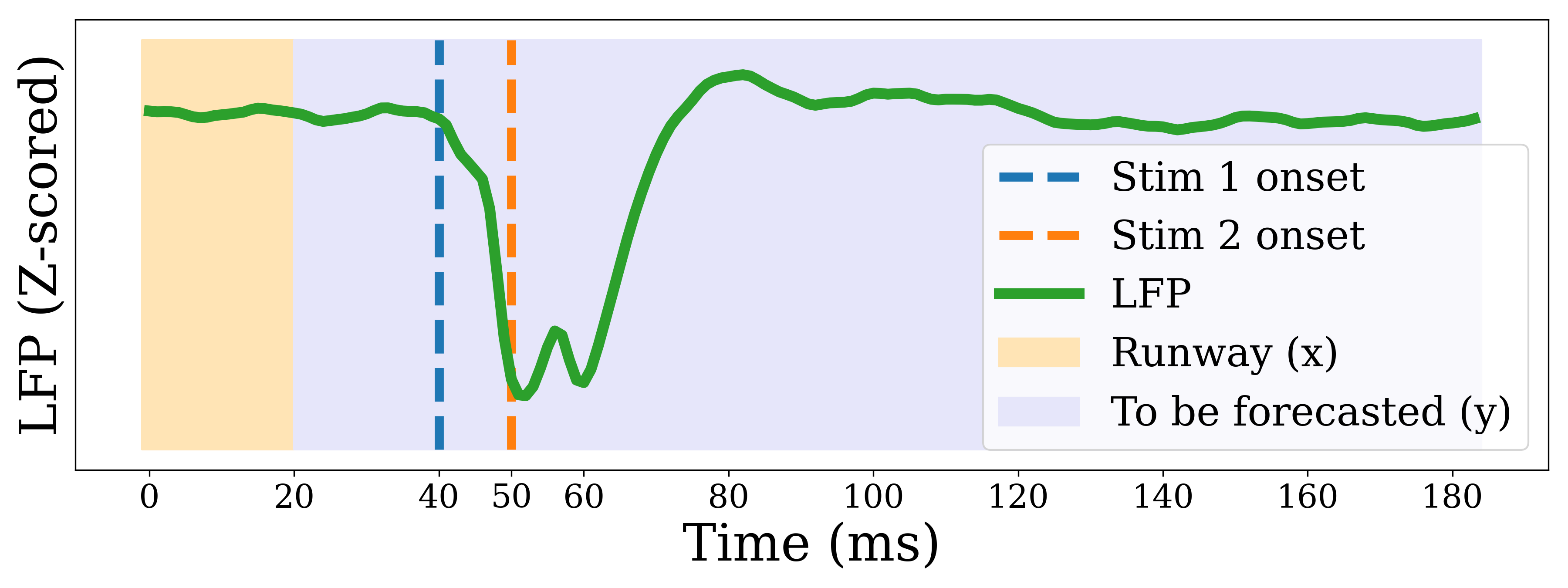}
  \caption{\textbf{Average response across many trials for a single electrode within one session.} The TBFM model predicts the stimulation-evoked activity ($y$) in each trial using the first 20 ms of recorded neural data ($x$, known as the ``runway''). The initial optical pulse occurs at 40 ms (blue dashed line); we assume there is a 20 ms control-loop delay between measuring the brain state at 20 ms and computing  parameters for the stimulation (see Sec.~\ref{subsec:kdg}). Note that the TBFM forecasts all channels simultaneously using the runway for all channels, but we depict only one channel here.}
  \label{fig:trial}
\end{figure}

We evaluate the proposed method using a previously published $\mu$ECoG dataset \cite{azadeh.data} comprising 40 optogenetic stimulation sessions collected from two rhesus macaques. In these experiments, excitatory neurons in left primary somatosensory (S1) and primary motor (M1) cortex were transduced to express the red-shifted channelrhodopsin variant C1V1. A semi-transparent 96-channel $\mu$ECoG grid enabled simultaneous recording of cortical local field potentials (LFPs) while paired optical pulses were delivered through two fiber-optic probes (Figs.~\ref{fig:spatiotemporal}–\ref{fig:trial}) \cite{azadeh.data2, azadeh.data3, azadeh.data4, azadeh.data5}. Each fiber optic was positioned adjacent to a specific ECoG electrode. Although the optical stimulation sites differed between sessions, they remained fixed within any given session. The number of electrodes meeting our quality criteria per session ranged from 42 to 94 (mean $78.6$, SD $13.2$), with electrodes being excluded based on impedance values measured during non-stimulation periods. Further descriptions of data acquisition and filtering procedures are provided in \cite{bloch.opto}.

To suppress line noise, we applied an infinite impulse response (IIR) notch filter at 60, 180, and 300~Hz, using quality factors (Q) of 20, 60, and 100, respectively. Results in this manuscript focus on the broadband time-domain signals; however, parallel analyses using bandpass- and time–frequency–filtered versions of the data are available upon request. Additional methodological details appear in \cite{azadeh.data, bloch.opto, bryan.jne}.

In each recording session, paired light pulses were delivered every $200ms$ in five consecutive 10-minute stimulation blocks, yielding approximately $15k$ pulse pairs per session. Individual pulses were $5ms$ in duration, and the inter-pulse interval within each pair was either $10ms$, $30ms$, or $100ms$ (session dependent). Stimulation blocks were interleaved with 10-minute rest periods during which neural activity was recorded without optical stimulation. During the experiments the animals were in an awake state watching cartoons.

\subsection{Introduction to temporal basis function models (TBFMs)}
This section provides a brief introduction to single-session TBFMs based on our early work \cite{bryan.embc}, and a significantly expanded analysis of TBFMs in our journal article \cite{bryan.jne}. A more detailed description as well as a reference $PyTorch$ implementation can be found here: \url{https://github.com/mmattb/py-tbfm}. This implementation is fit for use on a variety of stimulation modeling tasks, and is free to use under an open source license. The new extension of the single-session approach to multi-session is in Section \ref{sec:model}. 

\subsubsection{Key design goals}
\label{subsec:kdg}
We constructed the TBFM to solve several key challenges with machine learning approaches to neural stimulation modeling:
\begin{enumerate}
\item \textit{Sample efficiency}: Closed-loop experiments can collect only modest amounts of stimulation data due to safety and timing constraints; models must therefore learn effectively from small datasets.
\item \textit{Training efficiency}: Computationally expensive models and learning algorithms limit the types of adaptive protocols or rapid-iteration experiments we can run. Fast training enables broader experimental flexibility.
\item \textit{Low-latency inference}: Real-time stimulation requires predictions that remain valid within real-time control constraints. If model latency exceeds the forecasting horizon (e.g., a $50ms$ prediction with $75ms$ compute time), the controller acts on stale information and may not be able to effectively shape neural activity better than open-loop.
\end{enumerate}

\subsubsection{Architecture}
TBFMs provide single-trial, multi-step predictions of the neural response to stimulation. The model takes as input a short window of recently recorded neural activity—referred to as the ``runway'', together with the stimulation parameters. For this study, the runway consists of $20ms$ of neural data, and stimulation is assumed to be delivered $20ms$ later. This offset reflects an estimated worst-case control-loop latency, noting that some closed-loop systems achieve substantially lower delays (e.g., $7.5ms$ in \cite{guggenmos.latency}). During training, the TBFM is tasked with forecasting activity up to a horizon of 
t=$164ms$ (Fig.~\ref{fig:timeline}). In closed-loop demonstrations, only a part of this prediction window may be used, depending on the specific control objective \cite{bryan.embc}.

In our dataset, within a given session, stimulation was applied at fixed electrode sites with fixed stimulation parameters. As a result, the primary task of the controller trained on these data is to decide when to apply stimulation based on brain activity, rather than picking other parameters such as stimulation power or pulse width. Our pretrained TBFMs however must adapt to differing stimulation parameters because they vary across sessions. Specifically, the inter-pulse interval and target stimulation locations varied across sessions.

\begin{figure}[thpb]
  \centering
  \includegraphics[width=\linewidth]{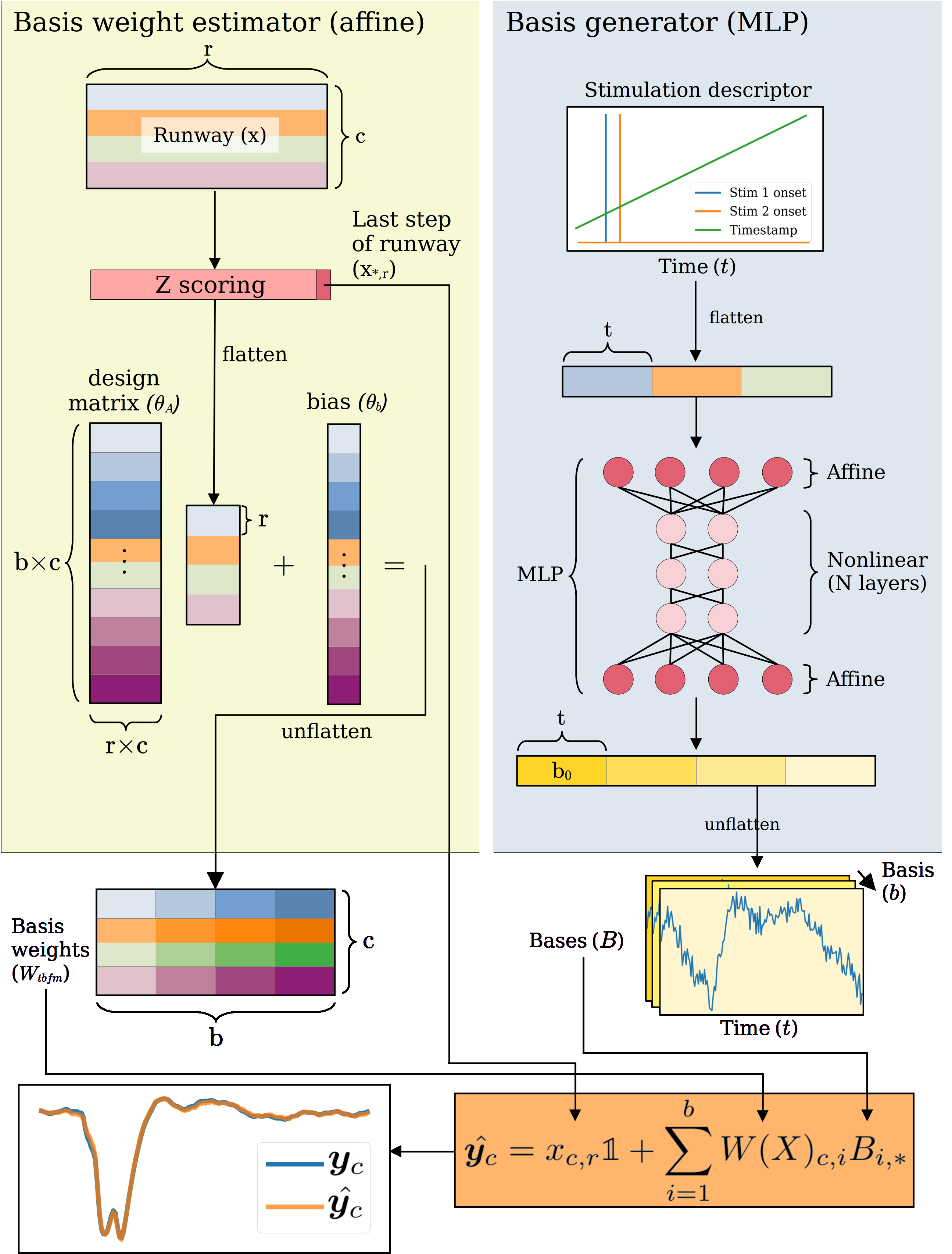}
  \caption{TBFM architecture}
  \label{fig:arch}
\end{figure}

Fig \ref{fig:arch} depicts the architecture of the TBFM (a more detailed description can be found in \cite{bryan.embc}). In brief, the single-session TBFM consists of:

\begin{enumerate}
    \item \textbf{A ``stimulation descriptor''} which is a tensor that encodes the stimulation parameters in some manner - their spatial pattern, timing, power, etc.
    \item \textbf{A basis vector generator} which receives as input any covariates useful for the forecast. In our stimulation setting, this includes the stimulation descriptor (see above). It outputs a set of basis vectors which the TBFM shares across all channels to make the forecast.
    \item \textbf{The ``runway''} as described above.
    \item \textbf{Channel-wise means and standard deviations used for Z-scoring the runway data}.
    \item \textbf{An affine basis weight estimator} which estimates the per-channel, time-invariant weights for each basis using the normalized runway as input.
    \item \textbf{The forecast}, defined as a weighted sum of the basis vectors and the last runway element for the given channel. That is, for a single channel $c$ the forecast becomes:
    $\hat{\boldsymbol{y}_c} = x_{c,r}\mathbf{1} + \sum_{i=1}^{b} W(X)_{c,i} B_{i,*}$, where $B$ is the matrix of basis vectors, $b$ is the basis count, $W$ is the weight matrix, $r$ is the runway length (see Fig \ref{fig:arch}).
\end{enumerate}

\subsubsection{Training}
\label{subsec:sstraining}
TBFMs are trained end-to-end with supervised backpropagation. For each channel, we estimate the mean and variance from the training set and apply Z-scoring using those fixed statistics during inference. The loss consists of the $L_2$ prediction error and a Frobenius-norm regularization on the basis weights.

Training uses mini-batches of $5k$ trials drawn from the early portion of each session, while evaluation is performed on the final 
$2.5k$ trials to assess robustness to within-session drift. This mimics a realistic deployment scenario where a model is calibrated early and applied later in the session. Optimization proceeds until the loss stabilizes, typically within $15k$ epochs.

\subsubsection{State dependency of predictions}
\label{subsec:statedep}
\begin{figure}
  \centering
  \includegraphics[width=\linewidth]{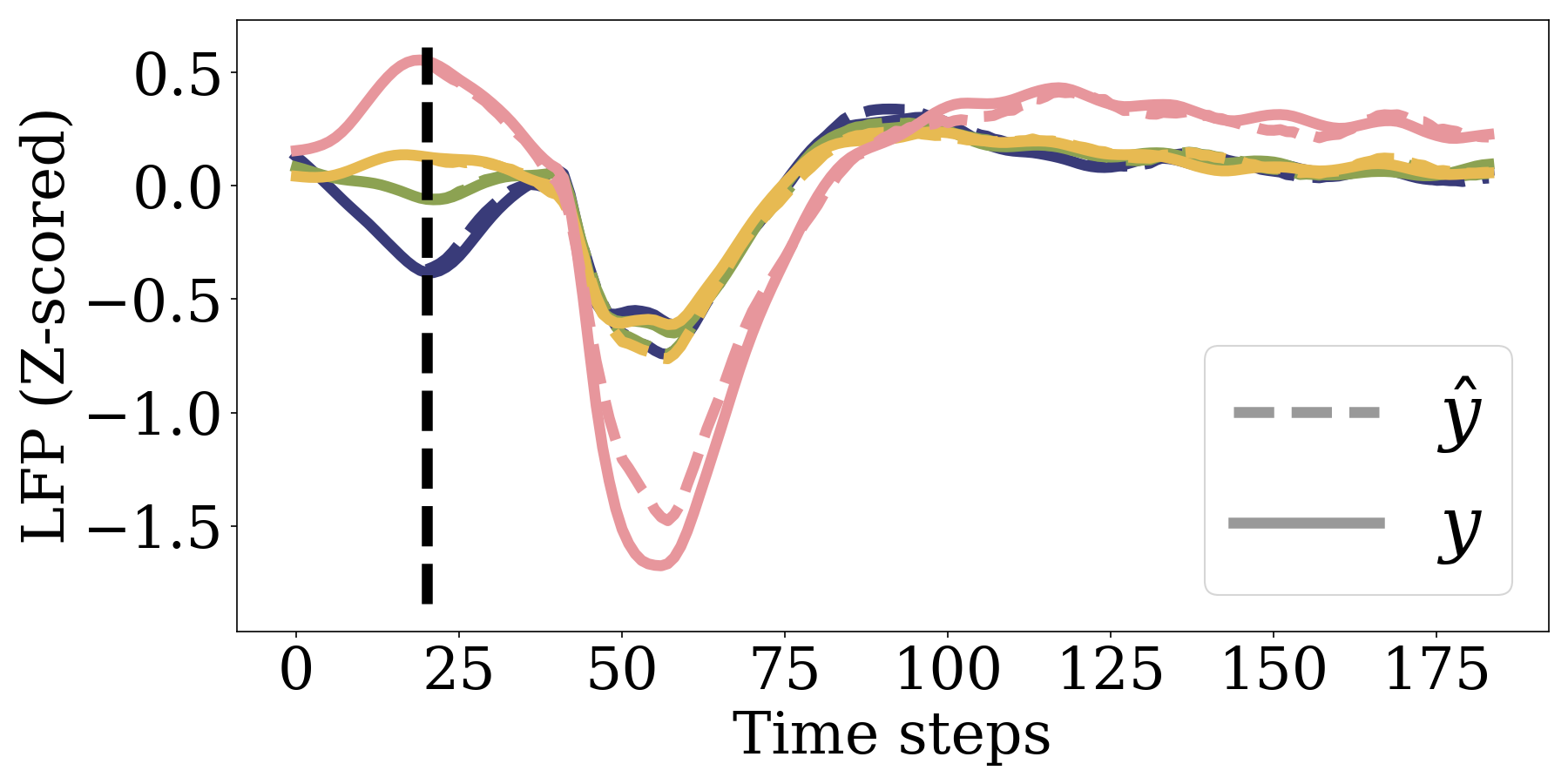}
  \caption{Mean neural response and mean prediction for four binned initial states. Single channel, single session. Stimulation onset is at t=$40ms$. $y$ is the mean stimulation response for trials in the given initial state bin, $\hat{y}$ is the mean forecast for the same trials.}
  \label{fig:statedep}
\end{figure}

Fig \ref{fig:statedep} depicts a TBFM's brain state-dependent predictions for a single channel and single session, where the TBFM was trained on that single session's data alone. For graphing, training set trials are binned into quartile ``initial states'' based on their value at the black dashed line, which is the last runway value. The pink ``high'' state exhibits a significantly different stimulation response than the others - one example of the brain state-dependency of the stimulation response present throughout these data. In \cite{bryan.jne} we present statistical evidence that this sort of state-dependency appears widely across the dataset and is statistically significant.

\subsubsection{Model compilation}
\label{subsec:compilation}
If stimulation parameters are drawn from a set of values - e.g. five different power levels - then an optimization is available to speed up inference time; we refer to it as ``compilation.'' Here the basis set for each parameter set is computed by forward pass through the basis generator, and saved in a dictionary. At inference time, the bases for that parameter set are pulled from the dictionary rather than computed on-the-fly, saving significant processing time, $\approx34\%$ for a typical session. We reuse this idea below.

\subsubsection{Performance comparison to linear state space models and recurrent neural networks}
In \cite{bryan.embc} we compared TBFMs to two other models which broadly represent common approaches to modeling neural dynamics, and more generally dynamical systems with control. The first was a simple linear state space model (LSSM) similar to one used previously for modeling neural responses to stimulation \cite{shanechi.stimmodel}. The other is a complex nonlinear recurrent neural network based on long short term memory (LSTM) cells \cite{murphy.ml}, which is strictly more expressive than the LSSM.

In brief: we found TBFMs to exhibit inference latency two orders of magnitude lower than either of the comparison models, while having the highest overall accuracy. TBFMs have sample efficiency slightly better than LSSMs and significantly better than the more complex LSTM model. As a result, TBFMs did not exhibit a trade-off between efficiency and accuracy on these data.

\subsubsection{TBFMs for closed-loop neural stimulation}
\label{sec:mpc}

Once we have built a TBFM we can specify a stimulation controller using approaches such as model predictive control (MPC) or model-based reinforcement learning (MBRL). In \cite{bryan.embc} we demonstrated TBFMs on two simulated closed-loop tasks: 1) applying stimulation only when key neural features are forecast to occur in the near future; and 2) timing stimulation to drive neural activity towards target dynamics. The first task reflects closed-loop strategies that deliver stimulation when specific neural states are detected, as in \cite{zanos.beta}, which triggers stimulation at particular beta-phase events, or \cite{castano.pd}, which adapts stimulation based on biomarkers of Parkinsonian symptoms. The second task parallels work in visual prostheses, where stimulation is used to drive neural activity toward patterns associated with particular visual inputs.

\subsection{A temporal basis function model (TBFM) with meta-learning}
\label{sec:model}
\begin{figure*}[thpb]
  \centering
  \includegraphics[width=\textwidth]{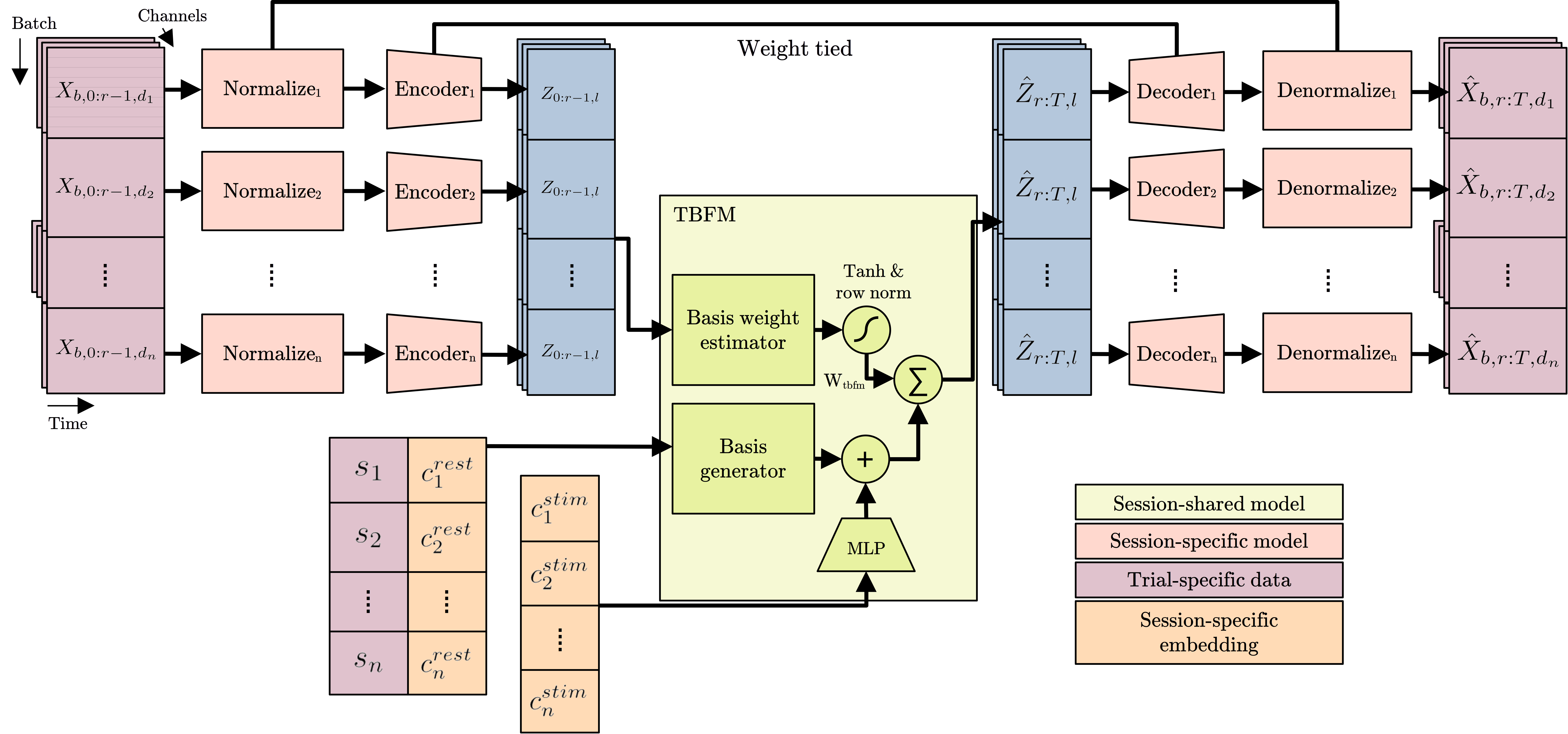}
  \caption{Multi-session TBFM architecture. $X_{b,t,d}$ Neural activity in a session (each block depicted is one of $n$ sessions), time span $t$, channel count $d_*$, multiple trials per session: $b$ is the batch dimension. $r$ is the runway length, $T$ is the forecast horizon. $Z_{t,l}$ Latent space representation of neural activity for a session, time span $t$, latent dimensionality $l$. $s_i$ per-trial embedding of stimulation parameters, known as the ``stimulation descriptor''. $c^{rest}_{s}$ per-session context from resting/non-stimulation activity, repeated for all trials of that session. $c^{stim}_{s}$ learned per-session stimulation context, repeated for all trials of that session.}
  \label{fig:csarch}
\end{figure*}

This paper builds on the single-session TBFMs by illustrating how they can be trained to generalize across many related experiment sessions, allowing for significant improvements in sample efficiency. The overall architecture is depicted visually in Fig \ref{fig:csarch}. We provide a table of ablations in \ref{subsec:ablations} to provide insight on the reasoning behind this design. The architecture extends the single-session model as follows:

\begin{enumerate}
    \item \textbf{Per-session normalizers} are not based on Z-scoring, but rather per-channel median and interquartile range (IQR) calculated from the training set of the given session. That is: the per-channel Z-score is replaced with: $\frac{X_{c} - Q_{50}}{Q_{75}-Q_{25}}$ where $Q_p$ indicates the $pth$ quantile of that channel's data from the training dataset. We found this to be marginally more robust than Z-scoring when working with small calibration data sets.
    \item \textbf{Per-session linear autoencoders} which map all sessions into a common latent space of dimensionality $l$. These provide two functions: 1) accounting for dimensionality differences due to differing numbers of missing channels between sessions; and 2) aligning the channels of the sessions in a latent space so that a common TBFM may be shared between them. We warm start the autoencoders with PCA, and a zero bias but refine them during training (see below). PCs are calculated from the session's training data.
    \item \textbf{A single session-shared TBFM} which lives in the latent space.
    \item \textbf{A basis weight normalization step} which 
    passes the basis weights through a hyperbolic tangent and $L_2$ normalization to force each channel's weight vector onto the unit sphere. This helps to stabilize training.
    \item \textbf{Per-session resting state context $c^{rest}_s$}. These context variables customize the bases generated by the TBFM to be more suitable for a particular session. The resting state context is constituted by autocorrelation statistics calculated on resting/baseline data gathered prior to stimulation being attempted. This is an optional covariate which informs the optimal basis shapes for the session (explained further in Section \ref{subsec:crest}).
    \item \textbf{Per-session stimulation context $c^{stim}_s$}. This vector is learned through our MAML-based optimization process (see below), and is passed into an MLP which reshapes the bases to customize them for the given session. The MLP outputs a low-rank matrix which is summed into the basis matrix - an approach known as low-rank adaptation (LoRA) \cite{lora}.
    \item \textbf{A training procedure (described below) based on model-agnostic meta-learning (MAML)} which optimizes a base TBFM to generalize well across sessions with minimal calibration data, while also optimizing the autoencoders and stimulation contexts $c^{stim}_s$.
    \item \textbf{A test-time adaptation (TTA) procedure (described below)} which fine tunes a pretrained TBFM for an unseen session using a small set of stimulation examples, known as the calibration data set. This is made efficient by the training procedure in the prior step, and is the primary way we take advantage of pretraining for unseen sessions.
\end{enumerate}

\subsubsection{Per-session resting state context}
\label{subsec:crest}
In a previous paper \cite{bryan.aacf}, we showed that autocorrelation statistics of baseline data were descriptive of the predictability of the \textit{stimulation data} from the same session. Specifically, the average absolute value of the autocorrelation function (A-ACF) calculated on baseline data correlated highly with the performance of all forecast model types we attempted on the stimulation data, including TBFMs. Autocorrelation is a proxy for the $R^2$ of a linear autoregressive forecast, and so it captures the (linear) predictability of a time series. It can also indirectly capture the signal-to-noise ratio (SnR) of the series under certain assumptions.

We leverage A-ACFs as covariates to help shape the temporal bases for a given session. While this is an optional aspect of our approach, it illustrates how additional sources of information could be leveraged to help shape a session's predictions. Conceptually, lower autocorrelations often mean that forecasts should be more conservative, and should revert quickly towards the channel mean. By passing A-ACF statistics to the TBFM, we can shape its bases to reflect a more conservative forecast if a channel or session have high noise.

For our case we measure A-ACF for each channel individually. $c^{rest}_s$ then becomes a vector containing the $25th$, $50th$ and $75th$ percentile A-ACF among all channels, i.e., the interquartile range of the A-ACFs. This context vector is calculated in advance on baseline data before any stimulation data is gathered.

\subsubsection{Per-session stimulation context}
\label{subsec:cstim}

An additional context vector $c^{stim}_s \in \mathbb{R}^{15}$ is optimized during training and during TTA to customize the TBFM's bases for that particular session. While sessions have similarities in their stimulation responses, the responses also have unique session-specific characteristics and bases need to be fine tuned accordingly.

The stimulation context vector is passed to an MLP which generates a low-rank correction summed into the matrix of bases. The correction is known as the ``residual matrix'', and this adaptation approach is referred to as ``low-rank adaptation (LoRA)'' \cite{lora}. The rank of this residual matrix is kept low so that it can be specified with a small number of parameters, which is necessary since it will be estimated using a very small dataset at test time. The basis matrix $B \in \mathbb{R}^{b,T}$ contains the $b$ bases, each of length $T$ (our forecast horizon). The residual matrix $R$ is the same shape, but estimated using a low rank factorization: $R = W O$ where $O \in \mathbb{R}^r$ is the correction of rank $r$ and output from the MLP, and $W \in \mathbb{R}^{b*T,r}$ is a matrix learned during training and frozen during TTA. The resulting bases then are computed as $B + R_{unflattened}$. Unflattening here refers to the reshaping of $R$ to match the shape of $B$. We used a residual rank $r$ of 16 for the results presented throughout this paper.

\subsubsection{Training}
\label{subsec:training}
The training algorithm, based on MAML, can be found in Algorithm \ref{alg:maml_training}. In brief, the training loop contains a type of simulation of the test-time adaptation procedure we will use later on unseen sessions. This causes the pretraining to push the model's cross-session parameters into a regime where test-time adaptation will yield an accurate model with a small number of training steps and a small calibration dataset. More specifically, the training procedure involves:
\begin{itemize}
    \item Random sample $n_{support}$ trials (usually 300) from the training batch, without replacement. These form the simulated ``support set'' - i.e. calibration set; the remaining trials are the ``query set''. This simulates the sampling process we will see when gathering an unseen session's support set, on which we will perform TTA.
    \item Train a newly-initialized stimulation context $c_s^{\text{stim}}$ until convergence using a simulated test-time adaptation procedure (TTA) and the support set. Training uses backpropagation and gradient descent. This is known as the ``inner loop''.
    \item Using the query set, calculate the loss (see below) and update the model parameters $\theta$ using backpropagation and a gradient descent based optimizer. This is known as the ``outer loop''.
\end{itemize}

This is a first-order MAML meta-learning algorithm which pushes the model parameters $\theta$ into a regime where they can be easily adapted through TTA. The random sampling in the inner loop ensures that the meta-learning objective is optimized over a distribution of possible support sets, rather than a fixed partition. This makes the learned embedding and adaptation procedure robust to the variability due to the calibration set sampling process we will see when we encounter an unseen session. By having a pretrained $\theta$ prior to TTA, we introduce a strong bias towards the use of cross-session patterns in the stimulation responses, which is likely the source of our robustness gains.

The inner loop loss function is based on $L_2$ forecast error $\| y - \hat{y} \|_2^2$. The outer loop loss function has the following terms:

\begin{itemize}
    \item \textbf{$L_2$ Forecast error} $\| y - \hat{y} \|_2^2$
    \item \textbf{Basis weight regularization} based on the Frobenius norm: $\lambda_{\text{weight}} \| W_{tbfm} \|_F^2$, where $W_{tbfm}$ is the TBFM's matrix of basis weights. This was found in \cite{bryan.embc} to reduce overfit on single-session TBFMs, and is retained here.
    \item \textbf{Reconstruction error} $\lambda_{\text{recon}} \| x - Dec(Enc(x)) \|_2^2$ which encourages the affine encoder and decoder to be true pseudo-inverses of each other.
    \item \textbf{A bases orthonormality penalty} $\lambda_{\text{ortho}} Ortho(B)$, 
    which encourages the bases to be both orthogonal to each other and unit length.
    Its mathematical definition can be found on our \href{https://mmattb.github.io/py-tbfm/#orthonormality-penalty}{Github pages}.
    \item \textbf{Stimulation context $L_2$ regularization} $ \lambda_{L2} \|c_s^{\text{stim}}\|^2$, which biases the stimulation context towards 0.
\end{itemize}

\begin{algorithm}
\caption{MAML-Based Multi-Session Training}
\label{alg:maml_training}
\begin{algorithmic}[1]
\REQUIRE Training data $\{\mathcal{D}_s\}_{s=1}^{S}$, model $f_\theta$ (TBFM + AE), config $\mathcal{C}$
\ENSURE Trained model parameters $\theta^*$
\STATE Initialize model parameters $\theta$, load rest embeddings $\{c_s^{\text{rest}}\}$
\FOR{epoch $= 1$ to $N_{\text{epochs}}$}
    \FOR{each minibatch $\mathcal{B} \subset \{\mathcal{D}_s\}$}
        \STATE $\mathcal{B}_{\text{support}}, \mathcal{B}_{\text{query}} \gets$ \textsc{RandomSplit}$(\mathcal{B}, n_{\text{support}})$
        
        \STATE \textbf{// Inner loop: adapt stimulus embeddings, simulating test-time adaptation (TTA)}
        \STATE Freeze $\theta$; initialize $\{c_s^{\text{stim}}\} \sim \mathcal{N}(0, 0.01)$
        \FOR{$t = 1$ to $T_{\text{inner}}$}
            \STATE $\mathcal{L}_{\text{inner}} \gets \frac{1}{|\mathcal{B}_{\text{support}}|} \sum_{s} \|\hat{y}_s - y_s\|^2 + \lambda_{L2} \|c_s^{\text{stim}}\|^2$
            \STATE $\{c_s^{\text{stim}}\} \gets \{c_s^{\text{stim}}\} - \alpha_{\text{inner}} \nabla_{c^{\text{stim}}} \mathcal{L}_{\text{inner}}$
        \ENDFOR
        \STATE $\{c_s^{\text{stim}}\} \gets \textsc{Detach}(\{c_s^{\text{stim}}\})$
        
        \STATE \textbf{// Outer loop: update shared parameters on query set}
        \STATE Compute $\mathcal{L}(\mathcal{B}_{\text{query}}, \theta, \{c_s^{\text{rest}}\}, \{c_s^{\text{stim}}\})$ \quad \textit{// see Loss function, Section \ref{subsec:training}}
        \STATE $\theta \gets \theta - \alpha_{\text{outer}} \nabla_{\theta} \mathcal{L}$
    \ENDFOR
\ENDFOR
\RETURN $\theta^*$
\end{algorithmic}
\end{algorithm}

\subsubsection{Test-time adaptation (TTA) procedure}
\label{subsec:tta}
Test-time adaptation (TTA) refers to the adaptation of a pretrained base model to an unseen task - in our case forecasting the effect of stimulation for an unseen session. It is based on a calibration data set, which is significantly smaller in size than what we would need for a single-session model.

The test-time adaptation (TTA) procedure reflects the same process as the training loop in Algorithm \ref{alg:maml_training}, but differs in some key ways:

\begin{itemize}
    \item The TTA support set is a small calibration data set we collected for the unseen session, rather than one we random sampled from a larger training set.
    \item The inner and outer loop structure remains the same, but only the autoencoder is fully trained in the outer loop. It is still warm started with PCA on the calibration dataset. The inner loop still optimizes $c^{stim}$. This is necessary since the autoencoder is specific to the session. The basis generator is fine tuned using a lower learning rate ($10^{-6}$).
\end{itemize}

\subsubsection{Multi-session model compilation}
\label{subsec:multi.compilation}
Like single-session TBFMs (Section \ref{subsec:compilation}), multi-session model compilation is possible. Compilation here refers to a simplification of the calculation which is possible after training and TTA have been performed. The simplification drastically reduces inference latency of the model (see Section \ref{sec:latency}).

Compilation in the cross-session case is performed after TTA, reducing the model to a moderately simple two layer neural network with a skip connection:

\begin{align}
\hat{\mathbf{y}}
=&
\Bigl(
  B^\top\,\phi\!\bigl(A_\text{pre}\,\text{vec}(\mathbf{x}) + v_\text{pre}\bigr)
  +\,\mathbf{1}_T\mathbf{z}_0^\top
\Bigr)\,W_{enc} \\
where: \\
\mathbf{z}_0 =& \mathbf{x}_r\,\tilde{W}_{enc}^\top + \tilde{b}_{enc} \\
\phi(P) =& rowNorm(tanh(P)) \\
\tilde{W}_{enc} =& W_{enc} \odot \boldsymbol{\alpha}, \\
\qquad
\tilde{b}_{enc} =& \boldsymbol{\beta}\,W_{enc}^\top + b_{enc}
\end{align}

Here $\alpha, \beta$ are the normalization constants used for the session's interquartile range (IQR) normalization, and $vec()$ vectorizes (flattens) a matrix. $A_\text{pre}$ and $v_\text{pre}$ fuse the normaliser, AE encoder, and basis-weighting layer into a single affine map: $\text{vec}(\mathbf{x}) \mapsto A_\text{pre}\,\text{vec}(\mathbf{x}) + v_\text{pre} \in \mathbb{R}^{l \cdot b}$, precomputed once post-TTA. The skip connection consists of the last runway value $\mathbf{x}_r \in \mathbb{R}^d$. Additional details on this derivation can be found on our \href{https://mmattb.github.io/py-tbfm/#multisession-model-compilation}{Github pages}.

Once we have adapted a model to a single session with TTA, we compile to reduce it to this simpler computation, resulting in a significant speed up in inference time (Section \ref{sec:latency}), and increasing the feasibility of deployment on embedded systems.

\subsubsection{Ablation: pretraining without MAML}
For comparison, we present results below for one additional training and TTA technique we call ``coadaptation.''  In this simpler pretraining method, MAML is not used, and instead $c^{stim}$ is optimized along with all other parameters (i.e. co-adapted). At test time as well, $c^{stim}$ is optimized jointly with the autoencoder. This model uses the same architecture but is trained and adapted with this simpler approach in order to measure the effect of MAML in our setting, relative to this simpler pretraining technique.

\subsubsection{Deployment}
\label{subsec:deployment}
Deployment refers to the process of building the base model over the course of a campaign of experiments and using it throughout the later part of the campaign to maximize robustness and minimize calibration data collection. There are many conceivable methods for deployment, and the optimal approach will likely depend on the details of the campaign. We present a simple deployment below to motivate future work.

Any deployment must address some key details. First: how much training data must we collect before we pretrain and use the base model? What decision rule do we use to go from single-session model use to the pretrained model and TTA use? For example: do we wait until the pretrained model has achieved a forecast error within a certain percentage of a single-session model on the same session? The answer to this question will depend on the accuracy/sample efficiency trade off we are willing to accept. A related question is, if enough compute is available, should we continue to train single session models on our tiny calibration set and compare their relative performance before deciding?
 
Second, will we continue to collect training data and refine the base model once we begin using it? How do we balance data collection for improving the base model and using it improve robustness and reduce data collection?

\begin{figure}
  \centering
  \includegraphics[width=\linewidth]{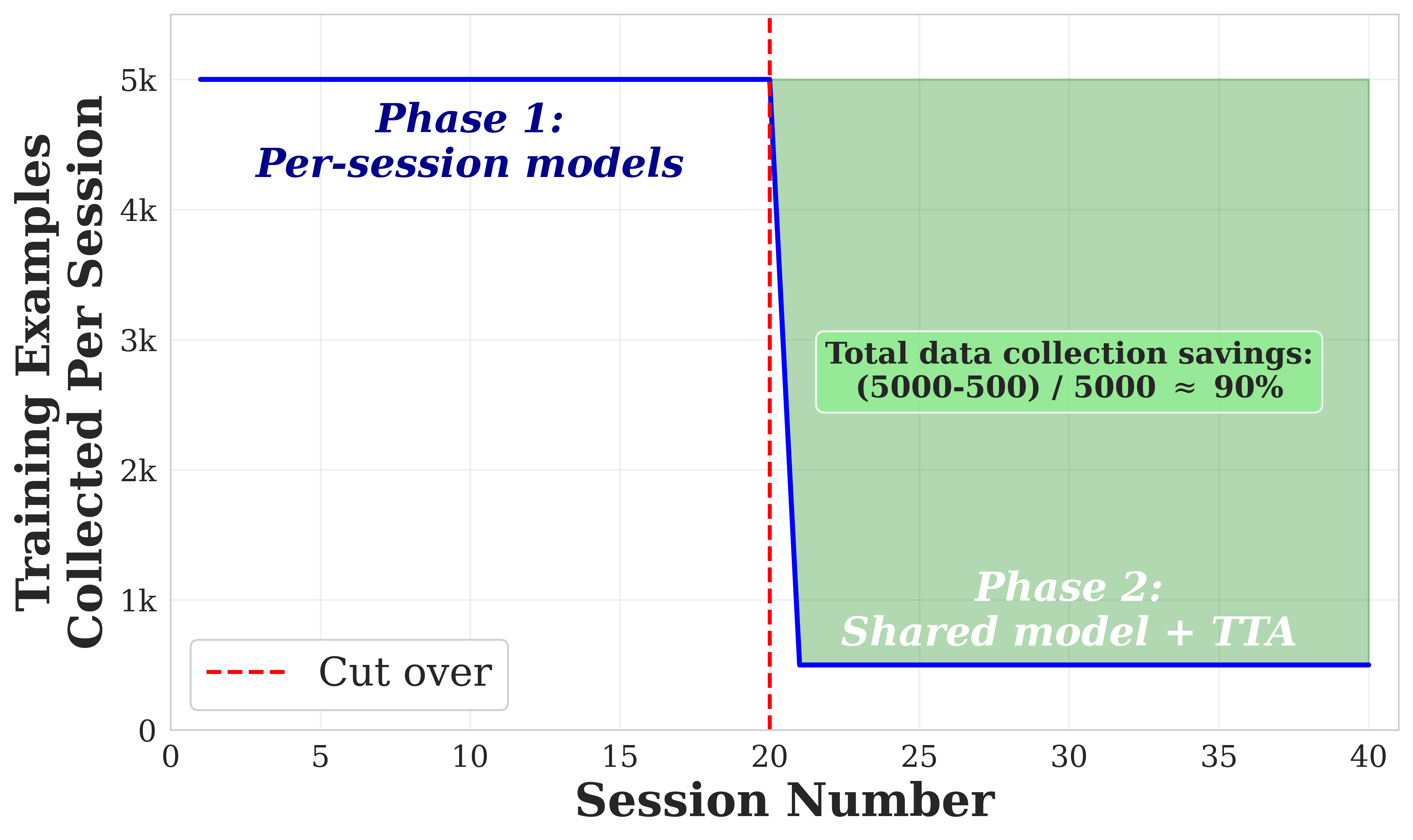}
  \caption{\textbf{Example deployment}. In this simple example we collect larger calibration (randomly timed stimulation) data sets during the initial 20 sessions. During those sessions we create a single-session TBFM for closed-loop experimentation. After 20 sessions, we use a meta-learned shared-session TBFM and test-time adaptation, allowing us to collect far smaller calibration data sets (5k versus 500 calibration examples).}
  \label{fig:flowhypo}
\end{figure}

Below we present results for a simple deployment, and leave more sophisticated data collection scheduling to future innovation. Our simple approach consists of collecting larger calibration data sets from the first N sessions of the campaign, and using single session models on those sessions. Once N sessions have passed, we pretrain the shared session model and deploy it for all future sessions. In these later sessions, we can collect much smaller calibration sets. This is graphically depicted in Fig \ref{fig:flowhypo}, with  the results described in section \ref{sec:results}.

\section{Results}
\label{sec:results}

For all results below, we evaluate our pretrained model types - co-adapted and MAML - on a difficult test time adaptation task. We hold out 20 sessions of data for  test-time adaptation (TTA), and pretrain on the remaining 20 sessions. We compare the accuracy of single-session models on the 20 held out sessions, using the same sized calibration data sets, allowing us to assess sample efficiency of the respective approaches. In all cases, we assess accuracy using a test set of $2.5k$ examples drawn from the end of the held out session, simulating the trials we would see later (typically 1 hour) after data calibration. The TTA is performed on trials from the first part of the session for the given calibration set size, once again simulating the flow of an experimental session. Rather than N-fold cross validation, we perform Monte Carlo Cross-Validation (MCCV). We repeat our process 20 times, each time assigning sessions randomly to the held-in/out portions. We use this in lieu of N-fold cross-validation to maintain sufficiently-sized held-in/out set proportions.

\subsection{Pretraining and MAML provide more robustness across sessions}
\label{subsec:robust}
A desirable property for stimulation models is robustness, meaning that their accuracy should degrade gracefully rather than failing catastrophically in the presence of stronger nuisance variables such as noise. Overall, we found that pretrained TBFMs exhibit significantly improved robustness across all calibration set sizes, defined as the mass of the left tail of the accuracy distribution across held-out sessions and MCCV splits.

\begin{figure}
  \centering
  \includegraphics[width=\linewidth]{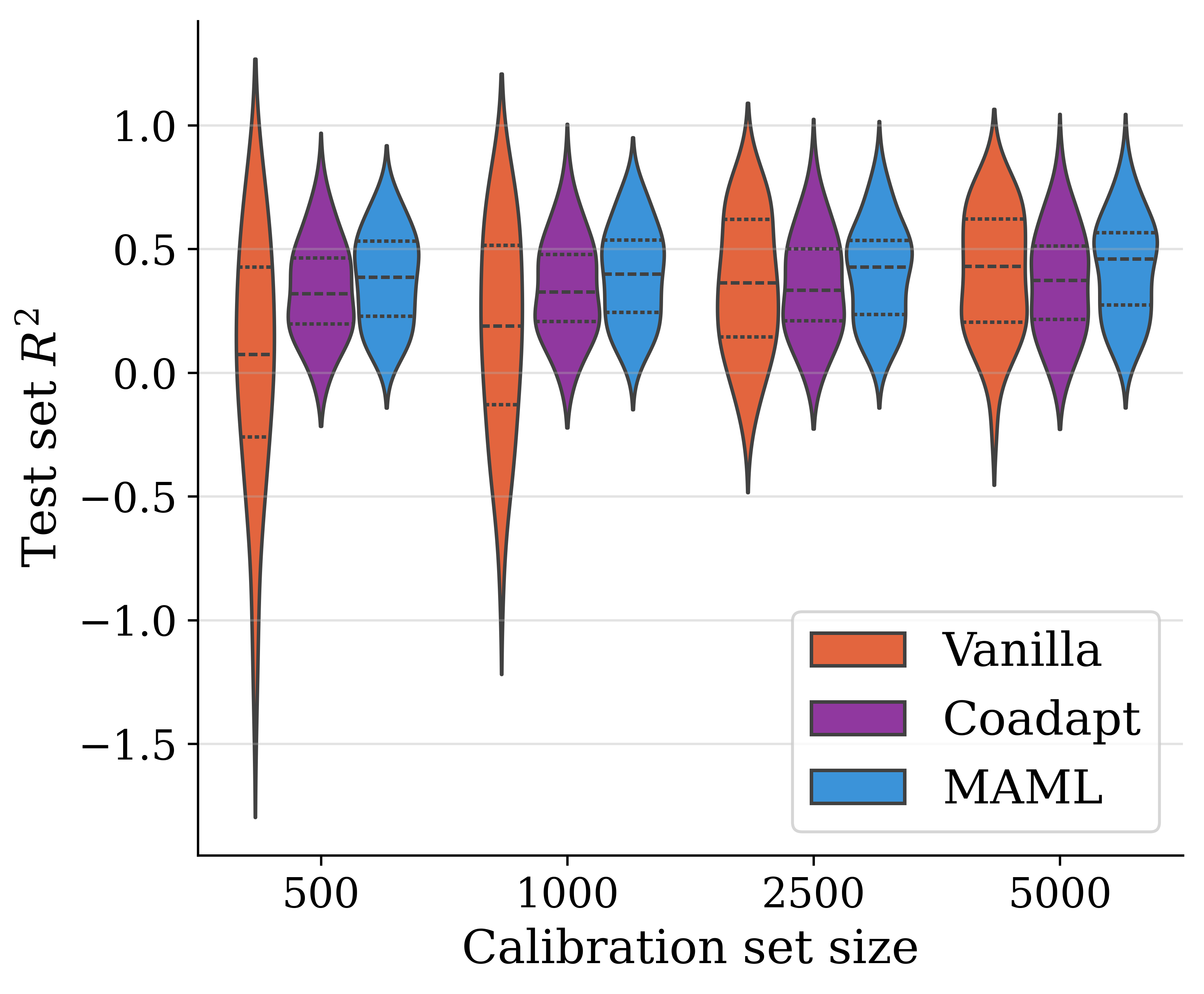}
  \caption{\textbf{Distribution of test set $R^2$ across folds and sessions, by calibration set size and training method.} Dashed lines indicate median and IQR. Left tails of the distributions are significantly heavier for vanilla (i.e., single-session) models, indicating greater robustness of the coadaptive- and MAML-pretrained models. At the calibration set size of $1k$ for example, 7 of 40 sessions with coadapted model, and 7 of 40 with MAML models have test set $R^2 < 0.15$, compared with 18 of 40 for vanilla models. At $2.5k$ it is 8, 6 and 11, respectively.}
  \label{fig:res.violin_compare}
\end{figure}

In Fig \ref{fig:res.violin_compare}, we see the distribution of test set $R^2$ across sessions for each calibration set size. In all cases, the left tail of the distribution is significantly longer for vanilla (single-session) models compared to the coadaptive and MAML approaches. At a calibration set size of $1k$ for example, 16 of 40 sessions have vanilla models with test set $R^2$ below $0.05$, compared with 1 of 40 for MAML-trained TBFMs. At a threshold of $R^2<0.15$ these numbers increase to 18 and 7, respectively. Measured in terms of variance, the variances at calibration set sizes $<5k$ is less for MAML models to a statistically significant degree under the single-sided Brown-Forsythe test at $p<0.05$, and $p = 0.15$ at the $5k$ set size.

\begin{figure}
  \centering
  \includegraphics[width=\linewidth]{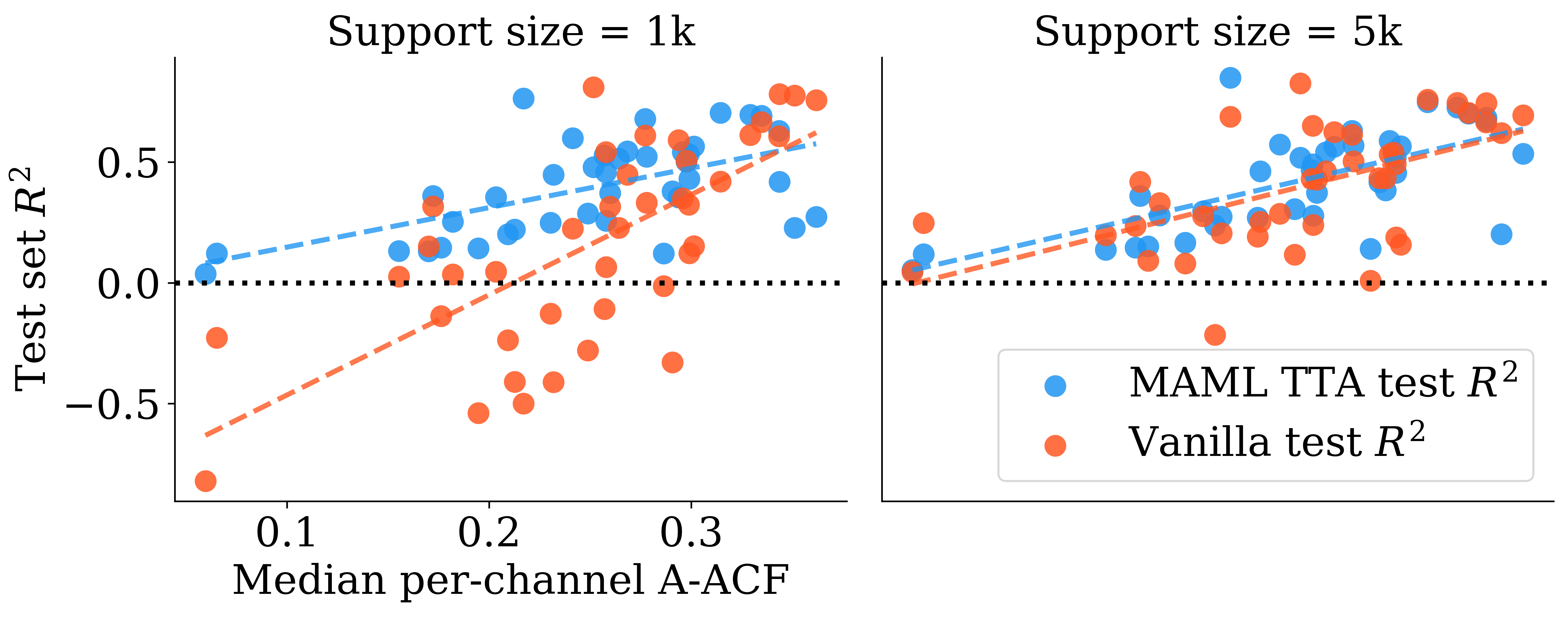}
  \caption{\textbf{A-ACF predicts performance of pretrained models. Robustness is indicated by high gains on noisy sessions.} Lower A-ACF values on resting data suggest a session's data are noisier. The median A-ACF value across a session's channels is predictive of both vanilla (single-session) model and MAML TTA model performance: the trend slope is positive at $p<5^{-5}$ in all cases. Pretraining flattens the susceptibility of the model to failing catastrophically on high noise (low A-ACF) sessions, and hence we see large improvements in the high noise regime. Vanilla models more frequently fail catastrophically ($R^2 < 0.05$) on higher noise sessions. At 5k calibration trials (right), the two methods converge, indicating that pretraining's advantage is specific to low-data regimes where anchoring to cross-session patterns compensates for limited per-session evidence. The slope of the difference in trends is negative at $1^{-3}$ for 1k, insignificant in 5k. The 5k result is expected since prior work found this was the optimal training set size for single-session models \cite{bryan.jne}.}
  \label{fig:res.acf}
\end{figure}

Fig \ref{fig:res.acf} further illustrates that we see the highest performance gains on high noise sessions (indicated by a low A-ACF value). A-ACF calculated on resting (i.e., non-stimulation) data can be a proxy for the noise in a session's data, and for the ability of a model to forecast that session's neural activity \cite{bryan.aacf}. We see in Fig \ref{fig:res.acf} that this relationship holds for pretrained and test-time adapted models, and the largest performance gains are on the noisiest sessions. 

We also see this robustness effect in Fig \ref{fig:res.support}, where the prediction interval (PI) for an unseen session test $R^2$ is significantly wider across all calibration set sizes for vanilla models than for MAML-trained models. Notably, we sometimes see vanilla model performance marginally exceed MAML models. At a $1k$ set size for example, 8 of 40 sessions had higher vanilla model performance. Luckily, since training times are so short (see Section \ref{sec:latency}), an experimenter is free to try both methods in a given session and pick the one which works best on some validation set.

\subsection{Pretraining and meta-learning significantly increase sample efficiency on unseen sessions}
Overall, our pretrained models significantly exceed in sample efficiency relative to our single session baseline. In all cases, we see diminishing returns on increasing calibration set size, but at drastically different rates between single session and pretrained models.

\begin{figure}
  \centering
  \includegraphics[width=\linewidth]{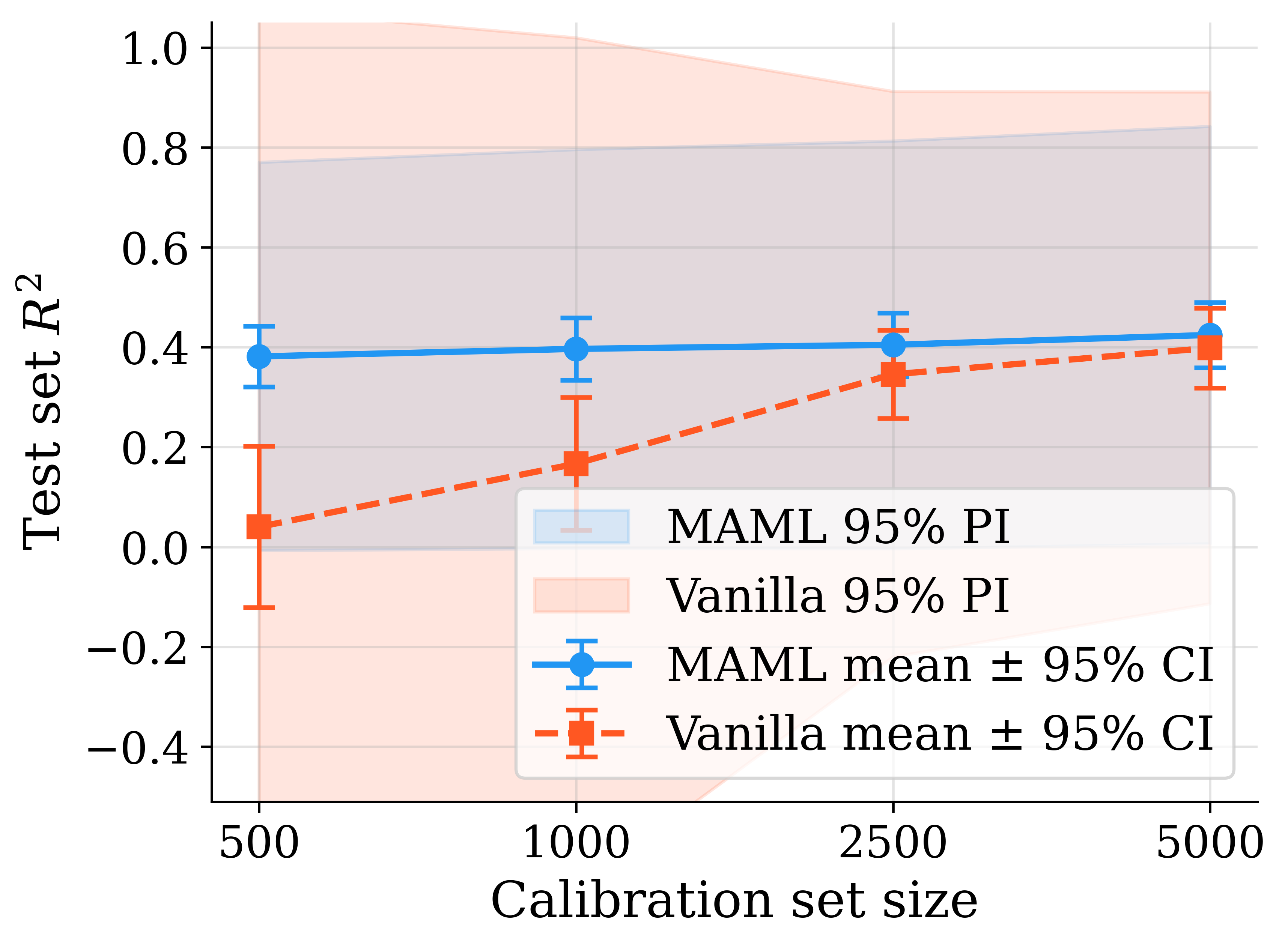}
  \caption{\textbf{Sample efficiency of single session
versus MAML pretrained models.} The plot shows mean and distribution of test set $R^2$ across held out sessions and Monte Carlo Cross-Validation (MCCV) splits. Confidence intervals (CIs) are for the mean of held-out sessions. The prediction intervals (PIs) capture the variability in performance across sessions and MCCV splits, reflecting both genuine cross-session differences in signal quality and the uncertainty introduced by the particular held-in/held-out assignment. Mean performances of MAML-pretrained models exceed vanilla (single session) models at $p<0.05$ under the 1-sided Wilcoxon signed-rank test for calibration set sizes $\le2.5k$, and $p=0.15$ for $5k$. Prediction intervals are significantly more narrow for MAML-pretrained models under the single-sided Brown-Forsythe test ($p<0.05$) for calibration set sizes $\le2.5k$, indicating more consistent performance and higher robustness. At $5k$, it drops to $p=0.19$.}
  \label{fig:res.support}
\end{figure}

At moderate calibration set sizes ($1k$-$2.5k$), the accuracy of the pretrained models exceed single-session models by a significant margin ($p<0.05$) on previously unseen (i.e., ``held-out'') sessions (Fig \ref{fig:res.support}). At a $1k$ calibration set size for example, single session models exhibit an average $R^2$ of $0.167$ on held out session test sets, while our MAML pretrained and adapted models exhibit an average $R^2$ of $0.397$ - a difference significant at $p=9.1^{-5}$ under the 1-sided Wilcoxon signed-rank test.

By a $5k$ set size, we see the performances of single session and pretrain models nearly converge, with pretrained models still showing a slight but statistically insignificant advantage: $0.424$ versus $0.398$ for single-session models - a difference $p=0.15$ under the 1-sided Wilcoxon signed-rank test. The convergence point represents the scale at which we have collected enough data that mean session performance begins to no longer benefit from pretraining. The experimenter saves on data collection by picking a calibration set size smaller than this. As we show in Section \ref{subsec:robust}, the pretraining approach additionally provides greater robustness, even at the point that the mean performances of the two approaches converge.

\subsection{MAML improves accuracy the over simpler pretraining approach}
MAML pretrained models exceed the co-adapted models in accuracy across all calibration set sizes (Fig \ref{fig:res.violin_compare}). At a $1k$ calibration set size, for example, we see $0.345$ $R^2$ for co-adapted models versus $0.397$ for MAML. At all set sizes MAML has significantly higher mean performance under the two-sided Wilcoxon signed-rank test ($p<1^{-5})$. These results indicate that meta-learning can help improve sample efficiency beyond mere pretraining.

The variance of results is also reduced by MAML across all calibration set sizes, though not to a statistically significant degree under the two-sided Brown-Forsythe test. We depict this visually in Fig \ref{fig:res.violin_compare}.  In brief: lower variance indicates more consistent results, and therefore may indicate a higher degree of robustness.

\subsection{Cross-animal generalization}
One desirable property of a cross-session stimulation model is the ability to generalize across subjects. Future foundation modeling efforts will rely on cross-subject similarity in the data. Our dataset contains sessions from two subjects: Subject G (22 sessions) and Subject J (18 sessions).

To examine cross-subject generalization, we performed an MCCV procedure similar to the results above. Here, for a given fold, we randomly split each subject's sessions into 10 held-in sessions, and the remainder are held-out. We pretrain each subject's model using MAML and its held in sessions, and then perform test-time adaptation of the model to the other animal's held out sessions. We repeat the TTA on the same-subject's held out sessions as well, to measure the difference between within-animal and cross-animal generalization. Due to computation limits, we perform only four folds, but note that the mean results varied little between folds.

Overall, we found that pretrained TBFMs successfully generalize between our two subjects. Results are summarized in Table \ref{table:cross-subject}. Held out sessions from Subject G exhibited test set accuracy effectively identical across models. Subject J's held out sessions yielded somewhat higher accuracy on the Subject J pretrained model, but the difference is not statistically significant under the two-sample t-test. These results indicate that 1) there exists enough similarity between subjects that a model pretrained on one subject can usefully adapt to an unseen subject; and 2) multi-session TBFMs are capable of performing this adaptation.

\begin{table}[]
\centering
\begin{tabular}{l|ll|}
\cline{2-3}
                                               & \multicolumn{2}{c|}{\textbf{Held out subject}}     \\ \hline
\multicolumn{1}{|l|}{\textbf{Held in subject}} & \multicolumn{1}{l|}{\textbf{G}}    & \textbf{J}    \\ \hline
\multicolumn{1}{|l|}{\textbf{G}}               & \multicolumn{1}{l|}{0.405 ± 0.052} & 0.354 ± 0.094 \\ \hline
\multicolumn{1}{|l|}{\textbf{J}}               & \multicolumn{1}{l|}{0.407 ± 0.047} & 0.373 ± 0.102 \\ \hline
\end{tabular}
\caption{Summary of Within- and Cross-Subject Generalization. Values are held-out session test set performance, mean across folds. Indicated interval is the $95\%$ confidence interval of the mean.}
\label{table:cross-subject}
\end{table}

\subsection{Training time and inference latency slower, but still faster than comparison models}
\label{sec:latency}
We benchmark training time and inference latency on a desktop computer built from common components: an AMD Ryzen Threadripper PRO 5955WX CPU, and an NVIDIA GeForce RTX 4090 GPU. In all cases training is performed on the GPU. We compare CPU versus GPU single-trial inference latencies in Table \ref{table:latency}.

Inference time for pretrained TBFMs is higher due to the higher model complexity, in particular the addition of an encoder/decoder. Typical GPU-based inference time for a single trial is $0.223ms$ for a pretrained and compiled TBFM, compared with $0.121ms$ for the compiled vanilla TBFM (see Table \ref{table:latency}). While slower, the pretrained TBFM remains well below our target $20ms$ real-time requirement, and faster than the latency of some existing clinical closed-loop systems at $7.5ms$ \cite{guggenmos.latency}. For comparison, the inference times were $10.111ms$ and $34.156ms$ for our baseline linear state space model (LSSM) and LSTM models, respectively.

\begin{table}[]
\centering
\begin{tabular}{l|r|r|}
\cline{2-3}
                                               & \multicolumn{1}{l|}{CPU} & \multicolumn{1}{l|}{GPU} \\ \hline
\multicolumn{1}{|l|}{Pretrained TBFM}          & 2.798                    & 0.790                    \\ \hline
\multicolumn{1}{|l|}{Pretrained TBFM compiled} & 2.095                    & 0.223                     \\ \hline
\multicolumn{1}{|l|}{Vanilla TBFM}             & 0.174                    & 0.226                    \\ \hline
\multicolumn{1}{|l|}{Vanilla TBFM compiled}    & 0.115                    & 0.121                    \\ \hline
\multicolumn{1}{|l|}{AE-LSTM}                  & 41.628                   & 34.156                   \\ \hline
\multicolumn{1}{|l|}{LSSM}                     & 10.111                   & 16.939                   \\ \hline
\end{tabular}
\caption{Average single-trial inference times (milliseconds)}
\label{table:latency}
\end{table}

Pretraining a multi-session TBFM on 20 held in sessions using MAML requires $3.6hr$ on average, which reflects the complexity of the training loop and scale of the training data set. We envision experimenters running this training between sessions rather than in the middle of a session, as needed.

During a session, the calibration data set can be collected, followed by TTA of a pretrained model or training of a vanilla model, followed by closed-loop stimulation. As a result, TTA and vanilla model training need to be fast. On our hardware, typical vanilla model training requires $\approx3 min$ for a calibration set size of $5k$. TTA requires $13min$ for an average session with a support size of $1k$, which is the support size with expected test set accuracy roughly equal to the vanilla model's dataset size of $5k$. While longer, this TTA training time is still well within typical experimental constraints. Compare to $5k$ training set times of $11.53hr$ and $90.6min$ for our complex LSTM and LSSM comparison models, respectively \cite{bryan.jne}.

Note that MAML with TTA in the typical sense is usually faster to train than a single session model: the intent is to initialize a network such that it can be trained with few gradient descent steps. Here MAML is playing a somewhat different role: it is used to initialize only the session stimulation embeddings ($c_s^{stim}$) and therefore temporal bases, not the entire network. This makes the more complex training loop (Algorithm \ref{alg:maml_training}) slower than the simple gradient descent used for single-session models. We believe this trade-off is justified: with higher accuracy, greater robustness, and reduced subject fatigue from calibration stimulation, we argue the increased training time is worthwhile, particularly given the promise of rapidly evolving hardware which can further accelerate the learning.

\subsection{Ablations}
\label{subsec:ablations}
Our model's design was based on the result of many ablations. These ablations provided insight into the design decisions needed to make the model work well for multi-session pretraining and adaptation. We believe these insights may generalize beyond our particular model and setting, and therefore outline the insights below. 

\begin{itemize}
    \item \textbf{Tanh nonlinearity and weight norm} were necessary for stabilizing training. Without these choices, we found that the basis weight estimator and basis generator parts of the TBFM network (Fig~\ref{fig:csarch}) have conflicting gradients, causing the learned parameters to oscillate in scale and never stabilize.
    \item \textbf{Basis orthonormality regularization} decreases held-in session accuracy but improves accuracy on held-out sessions, suggesting this regularization is a useful constraint which aids generalization.
    \item \textbf{IQR normalization rather than z-score} exhibits a lower median accuracy on held-out session test sets, across all support sizes. Its mean performance is higher at lower support sizes but to a statistically insignificant degree.
    \item \textbf{L2 regularization of $c^{stim}$} improves test set performance, especially on held-out sessions and lower support sizes. This penalty pushes basis adaptation to be moderate, and likely is controlling basis overfit.
    \item \textbf{Inclusion of $c^{rest}$} decreases held-in session performance, slightly decreases held-out session test set performance. We nevertheless include it in our results to illustrate the concept of leveraging exogenous correlates, as explained in Section \ref{sec:methods}.
    \item \textbf{MAML rather than co-adaptive pretraining} improves mean and median accuracy as well as robustness.
    \item \textbf{Autoencoder reconstruction loss} improved mean and median held-out session accuracies, but to a statistically insignificant degree.
    \item \textbf{Partial fine-tuning of latent TBFM during TTA}, allowing the basis generator to adapt at a low learning rate ($10^{-6}$) during TTA, helps accuracy on held-out session test sets, particularly at higher support sizes.
\end{itemize}

\section{Discussion}
\label{sec:discussion}

Our results show that pretrained temporal basis function models significantly exceed single-session models in sample efficiency and robustness, with MAML providing an additional and measurable improvement over simpler co-adaptive pretraining. Together our results provide, to our knowledge, the first demonstration that meta-learning can be applied to neural stimulation modeling to produce practical gains in efficiency and reliability.  We discuss below the interpretation of these findings, their architectural underpinnings, key limitations, and the longer-term trajectory toward foundation models for neural stimulation.

\subsection{Interpretation of Robustness Gains}
A finding at least as significant as the mean accuracy improvement is the compression of the left tail of the performance distribution (Fig~\ref{fig:res.violin_compare}). At a calibration set size of 1k, vanilla (single-session) models fall below $R^2 = 0.05$ - effectively no better than a forecast consisting only of the channel mean - on 16 of 40 sessions, compared with just 1 of 40 for MAML-pretrained models. In a clinical deployment, a system that achieves strong mean performance but fails entirely in a material fraction of sessions cannot be considered reliable. The prediction intervals (PIs) in Fig \ref{fig:res.support} quantify this: PIs reflect the variability in performance across sessions and MCCV splits, capturing both genuine cross-session differences in signal quality and uncertainty due to the particular held-in/held-out assignment. The significantly narrower PIs for MAML-pretrained models (Brown-Forsythe test, $p < 0.05$ for calibration set sizes $\leq 2.5$k) confirm that this is not merely a mean-shift effect.

We attribute the robustness improvement to two complementary mechanisms. First, the MAML objective explicitly optimizes for adaptation across a \emph{distribution} of support sets, sampled randomly during the inner loop. This discourages the model from finding parameter configurations that are sensitive to the particular trials included in the calibration set - precisely the configurations most likely to fail under noisy or atypical sessions. Second, the resting-state context $c_s^{\text{rest}}$, computed from autocorrelation statistics of baseline data prior to any stimulation, provides the model with an advance signal about session noise level. Our prior work showed that the average absolute value of the autocorrelation function (A-ACF) of baseline LFPs correlates strongly with the predictability of stimulation responses across model types \cite{bryan.aacf}. By conditioning the basis generator on this context, the model can modulate the conservatism of its forecasts before seeing a single stimulation trial, attenuating the catastrophic failures that arise when an overconfident model is deployed in a high-noise session.

\subsection{Interpretation of Sample Efficiency Gains}
Our core sample efficiency result, namely, that MAML-pretrained models achieve comparable or superior accuracy to single-session models using $50-90\%$ less calibration data, has direct practical consequences for closed-loop neurostimulation experiments. Calibration data collection is costly in at least three respects: it requires session time that could otherwise be devoted to closed-loop exploration, it imposes repeated non-therapeutic stimulation on the subject, and in clinical contexts, it may not be feasible at all within a single session. The convergence of pretrained and vanilla model performance at a 5k calibration set size ($R^2 = 0.424$ vs. $0.398$, $p = 0.19$) identifies the regime in which cross-session structure is no longer required for improving model accuracy, and single-session data alone is sufficient.  Below that threshold, the pretrained model exploits structure that the vanilla model must rediscover from scratch each session.

The diminishing returns observed for both model types as calibration set size increases are consistent with a general property of supervised learning: the most informative examples are encountered early, and additional data yields progressively smaller improvements in generalization. The pretrained model begins the calibration process in a $\theta$ neighborhood suitable for efficient adaptation. That initialization provides not only temporal bases capturing stereotyped stimulation-evoked response shapes, but also a weight estimator design matrix ($\theta_A$) encoding shared spatiotemporal predictive structure - specifically, the cross-session consistency in how pre-stimulus neural state in the runway predicts the evoked response. The stimulation context vector ($c_s^{stim}$) then captures session-specific deviations from these shared structures, requiring only a small support set to converge.


\subsection{The Role of Cross-Session Structure}
Meta-learning can succeed because the sessions share exploitable patterns. Local stimulation responses in primary somatosensory and motor cortex tend to be stereotyped at the level of individual channels: the canonical shape of a stimulation-evoked LFP response - an initial deflection followed by a rebound and recovery - is broadly conserved across animals and sessions \cite{bryan.jne}. This indirectly resembles the hemodynamic response function (HRF) modeled in functional magnetic resonance imaging (fMRI) studies, which is a temporal structure (i.e. temporal basis function) that can be leveraged across much of the brain in the context of a generalized linear model (GLM) \cite{friston.hrf}. Pretrained TBFMs likewise share temporal structure across both space and individual sessions, but learn the basis functions from data rather than using a canonical basis set. Inter-session variability reshapes the temporal bases, and the stimulation context $c_s^{\text{stim}}$ estimates the necessary reshaping in a low-rank manner. 

The per-session linear autoencoders further absorb differences in spatial structure. They serve two functions simultaneously: they handle the variable electrode count across sessions ($42-94$ usable channels in the present dataset), and they project each session's activity into a common latent space where the shared TBFM can operate. Warm-starting the autoencoders with PCA biases the encoders toward solutions that are compact. Together, these design choices explicitly factor out session-specific nuisance variation and expose the shared temporal structure that pretraining can leverage. This factored architecture may generalize naturally to other stimulation modalities and brain regions where comparable stereotypy exists.

\subsection{MAML vs.\ Co-Adaptive Pretraining}
The consistent accuracy advantage of MAML-pretrained models over co-adapted models ($p < 10^{-5}$ across all calibration set sizes) isolates the contribution of the meta-learning objective itself. The co-adapted model uses the identical architecture and training corpus, but optimizes $c_s^{\text{stim}}$ jointly with all other parameters rather than through a simulated inner-loop adaptation.  The fact that this simpler approach is already substantially better than a vanilla single-session model confirms that cross-session pretraining is broadly beneficial.  The further improvement from MAML suggests that the structured inner/outer loop training genuinely shapes the loss landscape in a way that facilitates adaptation, rather than merely benefiting from the larger effective training set.

Intuitively, MAML ensures that the stimulation embedding $c_s^{\text{stim}}$ is initialized in a region of parameter space where a small calibration set suffices to capture session-specific response properties. Without this, the co-adapted model may converge to representations that fit the training sessions well but do not support fast adaptation at test time - a form of overfitting to the meta-training distribution rather than to any individual session.  The residual-rank architecture (rank $r = 16$ for the basis residual matrix $R$) further supports this by keeping the degrees of freedom available to $c_s^{\text{stim}}$ small, ensuring that TTA remains a well-conditioned estimation problem even on tiny calibration sets.

\subsection{Limitations}

Several limitations of the present work warrant explicit discussion.

\subsubsection*{Dataset scope}
The dataset used in the present study derives from two rhesus macaques in a pair of cortical regions (S1/M1) using a single stimulation modality (optogenetic, red-shifted channelrhodopsin C1V1). While this allows careful characterization under controlled conditions, it limits the generalizability of our findings to other brain regions, species, and stimulation modalities. Electrical stimulation, for example, introduces stereotyped artifacts whose temporal shape may itself be well-captured by learned basis functions, but it remains unclear how to train a shared model on datasets that mix modalities with qualitatively different artifact time courses. Addressing this challenge - perhaps through modality-specific basis components or artifact-aware normalization - is an important direction for future work.

\subsubsection*{Linear autoencoders}
The choice of a linear autoencoder for cross-session alignment is deliberate, prioritizing computational simplicity and compatibility with small calibration sets.  Nonlinear alignment methods might achieve better generalization when sessions are more heterogeneous, but introduce additional parameters that are harder to warm-start and regularize with limited data.  The tradeoff between alignment expressivity and calibration efficiency deserves systematic study as larger, more heterogeneous stimulation datasets become available.

\subsubsection*{Test-time adaptation duration}
TTA times of approximately 13 minutes, while acceptable within a standard experimental session, represent a practical constraint in rapid-iteration paradigms and would need to be reduced for some clinical applications. The TTA runtime is dominated by autoencoder adaptation in the outer loop, not by the inner-loop stimulation context optimization.  Architectural modifications - such as a lower-dimensional autoencoder bottleneck, a stronger PCA-based initialization, or frozen encoder weights with only the decoder adapted - may
reduce this substantially. Continued improvements in GPU hardware will also provide gains independently of architectural changes.

\subsubsection*{Evaluation under distribution shift due to behavioral states}
Our evaluation protocol trains on the early portion of each session and evaluates on the late portion (final $2.5k$ trials), providing some protection against within-session drift. However, we have not evaluated performance under intentional distribution shift - for example, calibrating in one behavioral state and deploying in another. While known behavioral states may be captured as additional covariates that can be input to our basis weight generator, it remains unclear how well our model will generalize to unseen behavioral states. In clinical applications such as adaptive DBS, stimulation may be required across a wide range of patient states (rest, movement, sleep, emotional arousal, etc.), each potentially producing different response dynamics. Building explicit robustness to state-level distribution shifts into the meta-learning framework, perhaps through multi-state calibration protocols or explicit behavioral state conditioning, remains important future work for our approach and other stimulation modeling approaches.


\subsection{Active learning and adaptive deployment}
The deployment scenario described in Section \ref{subsec:deployment} is deliberately simple: collect full calibration sets for an initial batch of sessions, then switch to TTA for all subsequent sessions.  Myriad more sophisticated deployment strategies are conceivable. Active learning \cite{murphy.ml} is a particularly promising direction: rather than collecting calibration trials at random stimulation timings, the model could select the stimulation parameters or timing expected to maximally reduce its uncertainty about the session-specific response. In the closed-loop stimulation setting, this could take the form of an exploration phase - brief, model-guided stimulation trials chosen to be maximally informative - followed by a shorter exploitation phase using closed-loop control. Active learning has been shown to improve sample efficiency in related neural decoding settings, and its application to stimulation modeling is a natural next step.

Continuous online adaptation is another avenue: rather than treating TTA as a one-time calibration step, the model could be updated incrementally as new trials accumulate within a session, potentially tracking within-session non-stationarities such as habituation, electrode drift, or changes in arousal. A principled combination of online adaptation and active trial selection could substantially reduce the total number of stimulation trials required for reliable closed-loop control.

\subsection{Towards a foundation model for neural stimulation}
The results presented here provide, to our knowledge, the first empirical evidence that a pretrained model based on meta-learning can exploit cross-session structure in neural stimulation data to improve generalization in stimulation response forecasting. This is a necessary condition - though not sufficient on its own - for any credible effort to build foundation models for neural stimulation.

Large pretrained models have recently begun to emerge in neural decoding, driven in part by efforts like the FALCON benchmark \cite{falcon}, which has standardized evaluation of few-shot decoding across multiple recording modalities and participants. Stimulation modeling has lagged behind decoding in this and many regards, partly because ground-truth stimulation-response data is harder to collect at scale. The present approach begins to address this gap by demonstrating that pretraining is practical with modest multi-session datasets (20 sessions here), and that even limited cross-session structure is sufficient for meaningful improvement.

A practical path toward a stimulation foundation model based on TBFMs might proceed as follows. First, assemble a heterogeneous corpus of stimulation experiments spanning modalities, brain regions, and species. Second, use the TBFM architecture proposed here - with flexible per-session autoencoders, stimulation context vectors, and resting-state context - to learn a shared prior over stimulation responses. Third, evaluate generalization to held-out sessions under the full range of cross-session variability present in the corpus, using a benchmark structured analogously to FALCON. The resting-state context already illustrates one avenue for including external covariates that do not require stimulation trials; further biological covariates (electrode impedance profiles, resting-state power spectra,
subject demographics) could similarly inform adaptation without adding to calibration data requirements.

The key open question is whether the shared latent structure of stimulation responses is sufficiently consistent across the substantial biological heterogeneity of such a corpus to support a common prior. We consider this an important open question for the field, and call on experimenters to consider how standardized multi-site stimulation datasets - analogous to FALCON but for stimulation rather than decoding - could be assembled and shared to enable rigorous benchmarking. The present results provide early grounds for optimism that such an effort would be scientifically productive.


\subsection{Clinical translation considerations}
Beyond sample efficiency and accuracy, a credible path toward clinical translation requires predictable behavior under the full range of conditions likely to be encountered in use. Our robustness results are encouraging in this respect: the reduction in catastrophic failure modes suggests that MAML-pretrained models are less sensitive to variability in the calibration data, which is one proxy for the kind of distribution shift that clinical systems routinely encounter.

Several additional considerations will need to be addressed as this work moves toward clinical evaluation. Regulatory requirements for closed-loop neural devices generally demand that stimulation controllers behave safely and predictably under both normal and degraded operating conditions: a model that performs well on average but fails unpredictably in a minority of sessions is unlikely to meet this bar. The improved tail behavior we observe is a step in the right direction, but formal uncertainty quantification - for example, conformal prediction bounds on the forecast error - would provide stronger guarantees suitable for safety-critical deployment.

Real-world closed-loop systems also operate under hardware and energy constraints that laboratory systems do not face. The inference latency of $0.223ms$ for a pretrained TBFM on GPU is well within our $20ms$ target and faster than some
existing clinical systems \cite{guggenmos.latency}. The architectural simplicity of TBFMs makes them amenable to implementation on low-power embedded processors, a practical advantage over recurrent neural networks or transformer-based alternatives. Despite the training-time complexity, the post-TTA compiled model reduces to a shallow computation: two affine transformations separated by a normalized tanh nonlinearity, plus a skip connection carrying the last runway value forward as a baseline forecast. The training-time apparatus - per-session autoencoders, MAML inner loops, MLP-generated LoRA-based basis adaptation - collapses at inference into a structure resembling a two layer perceptron with a residual connection. This separation between training-time complexity and inference-time simplicity is architecturally favorable for clinical deployment: the elaborate machinery needed to learn cross-session structure does not impose corresponding complexity on the deployed device.

Finally, the inter-session variability we observe across just two animals underscores the challenge of building systems that generalize across patients with different anatomies, disease states, and stimulation histories. Transfer learning approaches like the one demonstrated here will likely be a prerequisite for practical deployment, given that collecting large per-patient calibration datasets is often infeasible.  We view the results reported here as an early proof of concept for this broader program.

\section{Conclusion}
\label{sec:conclusion}
Our results provide, to our knowledge, the first demonstration that meta-learning can be applied to neural stimulation modeling to achieve meaningful improvements in both sample efficiency and robustness. Building on temporal basis function models, we show that model-agnostic meta-learning (MAML) enables a pretrained model to adapt to unseen sessions using calibration datasets $50-90\%$ smaller than those required by single-session models, without a significant accuracy trade-off at moderate calibration set sizes. Critically, pretraining does not merely shift mean performance upward - it substantially compresses the left tail of the performance distribution, reducing the frequency of catastrophic failures that would otherwise make deployment unreliable.

The architectural foundation for these gains is a factored design in which shared temporal basis functions and a shared weight estimator design matrix capture cross-session structure, while per-session autoencoders and stimulation context vectors absorb session-specific variation. This division of labor reflects a broader principle with precedent in functional neuroimaging, where stereotyped stimulation response timecourses - analogous to the hemodynamic response function - have long been leveraged to improve signal estimation across brain regions and subjects \cite{friston.hrf}. Our results suggest that a similar principle holds for LFP responses evoked by neural stimulation applied directly to neural tissue (as opposed to, e.g., visual stimulation applied through the eyes). Our approach additionally has the advantage that the basis functions are learned from data rather than assumed canonical.

Our approach simultaneously yields inference latency well within real-time control constraints, and the MAML training objective provides robustness properties that begin to address the reliability requirements for clinical deployment. These results provide early empirical evidence that cross-session structure in neural stimulation data is consistent enough to support pretraining - a necessary condition for any future effort to build foundation models for neural stimulation.

Taken together, we believe this work meaningfully narrows the gap between complex data-driven approaches to closed-loop neural stimulation and the practical constraints - limited calibration time, session-to-session variability, and noise - that have historically made such approaches difficult to deploy. The path forward involves extending this framework to more heterogeneous datasets spanning modalities, brain regions, and species, and developing standardized benchmarks that can drive progress toward a general-purpose neural stimulation foundation model.

\addtolength{\textheight}{-12cm}   


\addtolength{\textheight}{+12cm}
\FloatBarrier
\newpage
\vspace{1pt}
\clearpage
\section{Appendix}
\subsection{Code availability}
All code for this paper is available through an open-source license (BSD 3-Clause) in \href{https://github.com/mmattb/py-tbfm/tree/multisession}{the `multisession' branch of our Github repository}.

\subsection{Hyperparameters}
\label{apx:hyperparameters}

The reader can find most of these hyperparameters defined in our \href{https://github.com/mmattb/py-tbfm/tree/multisession/conf}{configuration files},
allowing them to trace the values easily through the code.

\addtolength{\textheight}{+12cm}

\subsubsection{Model architecture}
\begin{center}
\footnotesize
\begin{tabular}{@{}ll@{}}
\toprule
Hyperparameter & Value \\
\midrule
\multicolumn{2}{@{}l}{\textit{TBFM}} \\[2pt]
Ambient latent dimension ($l$) & $96$ \\
Covariate (stim) dimension & $3$ \\
Number of bases ($b$) & $100$ \\
Basis-generator latent dim & $16$ \\
Basis-generator depth & $3$ \\
Basis-generator dropout & $0.3$ \\
Meta projection dim & $32$ \\
Rest-embedding dim ($c^{\text{rest}}_s$) & $3$ \\
Stim-embedding dim ($c^{\text{stim}}_s$) & $15$ \\
$\tanh$ basis weights & true \\
Basis-weight row norm & true \\
Shared bases & true \\
[4pt]\multicolumn{2}{@{}l}{\textit{Meta-learning residual module}} \\[2pt]
Basis residual rank & $16$ \\
Residual MLP hidden dim & $16$ \\
[4pt]\multicolumn{2}{@{}l}{\textit{Autoencoder}} \\[2pt]
Autoencoder (AE) type & \texttt{LinearChannelAE} \\
AE input dim & $96$ \\
AE bias & true \\
\bottomrule
\end{tabular}
\end{center}

\newpage
\subsubsection{Training}
\nopagebreak[4]
\begin{center}
\footnotesize
\begin{tabular}{@{}ll@{}}
\toprule
Hyperparameter & Value \\
\midrule
\multicolumn{2}{@{}l}{\textit{Data}} \\[2pt]
Trial length & $184$ bins \\
Runway (pre-stimulus conditioning window) & $20$ bins \\
Prediction length / forecast horizon ($T$) & $164$ bins \\
Train set size & $5000$ \\
Test set size & $2500$ \\
Held-out sessions & $15$ \\
Normalizer & quantile \\
Outlier IQR multiplier & $10.0$ \\
Max outliers per trial$^a$ & $5$ \\
[4pt]\multicolumn{2}{@{}l}{\textit{Outer loop}} \\[2pt]
Training steps ($N_{\text{epochs}}$) & $12001$ \\
Batch size per session & $500$ \\
Gradient clip ($\ell_2$) & $2.0$ \\
Basis weight estimator updates per basis generator update$^b$ & $3$ \\
Basis weight estimator LR & $4\times10^{-4}$ \\
Basis generator (outer) LR ($\alpha_{\text{outer}}$) & $3\times10^{-3}$ \\
Basis weight estimator weight decay & $1\times10^{-4}$ \\
Basis generator weight decay & $1\times10^{-5}$ \\
Basis weight reg.\ weight ($\lambda_{\text{weight}}$) & $75.0$ \\
Orthonormality penalty ($\lambda_{\text{ortho}}$) & $0.05$ \\
AE reconstruction weight ($\lambda_{\text{recon}}$) & $0.03$ \\
AE co-adaptation (fine-tune after PCA warm-start) & true \\
AE freeze step (step at which AE is frozen) & $5000$ \\
AE LR & $1\times10^{-4}$ \\
AE optimizer & Adam ($\varepsilon=10^{-8}$, \\
&AMSGrad) \\
Normalizer LR & $1\times10^{-4}$ \\
[4pt]\multicolumn{2}{@{}l}{\textit{Inner loop}} \\[2pt]
Inner-loop steps ($T_{\text{inner}}$) & $20$ \\
Tail inner steps$^c$ & $1000$ \\
Support set size ($n_{\text{support}}$) & $300$ \\
Stim-embedding $\ell_2$ weight ($\lambda_{L2}$) & $5\times10^{-4}$ \\
Inner LR ($\alpha_{\text{inner}}$) & $1\times10^{-2}$ \\
Inner weight decay & $1\times10^{-4}$ \\
\bottomrule
\multicolumn{2}{@{}p{0.85\columnwidth}}{\scriptsize
$^a$~A trial is excluded if more than this many timepoints are flagged as
outliers by the IQR criterion.
$^b$~Decouples the optimization rates of the two parameter groups.
$^c$~Additional inner-loop steps applied to the stim embedding after the final outer
TTA loop, allowing a re-fit on the full support set.} \\
\end{tabular}
\end{center}

\subsubsection{Test-time adaptation (TTA)}
\nopagebreak[4]
\begin{center}
\footnotesize
\begin{tabular}{@{}ll@{}}
\toprule
Hyperparameter & Value \\
\midrule
TTA steps & $7001$ \\
Support sizes swept & $\{500, 1000, 2500, 5000\}$ \\
Max adapted sessions & $20$ \\
Progressive-unfreezing threshold$^d$ & $0$ (always) \\
Fine-tune basis generator (\texttt{--unfreeze-bases}) & yes \\
Basis-generator LR & $1\times10^{-6}$ \\
Fine-tune basis weights & no \\
Basis-weight LR (when unfrozen) & $1\times10^{-5}$ \\
\bottomrule
\multicolumn{2}{@{}p{0.85\columnwidth}}{\scriptsize
$^d$~Basis generator fine-tuning is enabled only when the support size exceeds
this value; 0 means always enabled.} \\
\end{tabular}
\end{center}

\end{document}